\documentclass[runningheads]{llncs}

\usepackage{eccv}

\usepackage{eccvabbrv}

\usepackage{graphicx}
\usepackage{booktabs}
\usepackage{colortbl}
\usepackage{subcaption}

\usepackage[accsupp]{axessibility}  

\usepackage{fontawesome5}
\usepackage{multirow}
\usepackage{makecell}
\usepackage{array}
\usepackage{algorithm}
\usepackage{algpseudocode}

\usepackage[hypertexnames=false,breaklinks]{hyperref}
\hypersetup{colorlinks=true, linkcolor=eccvblue, citecolor=eccvblue, urlcolor=eccvblue}

\usepackage{orcidlink}

\begin{document}

\title{MoE-ViE: Mixture of Experts Vision Encoder for Efficient Image and Video Understanding}

\titlerunning{MoE-ViE}

\author{Bonan Zhang\inst{1} \and
Shiyu Dong\inst{1} \and
Quan Hung Tran\inst{1} \and 
Katharina Gschwind\inst{\star} \and
Shuqi Yang\inst{\star} \and
Sijia Chen\inst{\star} \and
Adel Ahmadyan\inst{1} \and
Seungwhan Moon\inst{1} \and
Lu Zhang\inst{1} \and
Ahmed Kirmani\inst{1} \and
Babak Damavandi\inst{1} \and 
Anuj Kumar\inst{1} 
}

\authorrunning{B.~Zhang et al.}

\institute{$^{1}$Meta\\
$^{\star}$Work done while at Meta\\
\email{\{bonanz,shiyudong,quanhungtran,kgschwind,shuqiyang
,sijiac,ahmadyan,luzhang20,kirmani,babakd,anujk\}@meta.com}}

\maketitle

\begin{abstract}
Vision encoders are a critical component of vision-language models, and scaling their capacity effectively improves performance. However, dense scaling increases compute cost and inference latency. Mixture-of-Experts (MoE) architectures offer a compelling alternative, having enabled efficient scaling in LLMs, yet the MoE design space for CLIP-style vision encoders remains underexplored at State-of-the-Art (SOTA) levels. In this work, we systematically study MoE designs for vision encoder scaling and find that fine-grained MoE topologies yield substantial gains over both dense and standard MoE counterparts. We further propose an auxiliary-loss-free balancing variant for better expert utilization, and design a specialized MoE kernel to mitigate inference latency overhead. To enhance video capabilities while preserving image knowledge, we introduce frame-level distillation paired with a novel freezing mechanism. We pretrain a series of Mixture-of-Experts Vision Encoders (MoE-ViE) across a range of sizes, all consistently outperforming their dense counterparts. Our largest model matches the zero-shot performance of a SOTA encoder $1.7\times$ its size at $76\%$ of its latency. When aligned with an LLM, MoE-ViE surpasses all compared encoders on image and video benchmarks, including those with up to $5\times$ more activated parameters. Code is available at \url{https://github.com/facebookresearch/moe_vie}.

  \keywords{Vision encoder \and Mixture-of-Experts \and CLIP}
\end{abstract}

\section{Introduction}
\label{sec:intro}

Vision encoders have become a cornerstone in today's vision-language models (VLMs). The pursuit of higher accuracy and broader applicability necessitates the continuous scaling up of vision models, which in turn has been a reliable route to stronger VLM capabilities \cite{pe,internvl2p5,imagebind,eva-clip-18b}. However, scaling dense vision encoders comes with system-level trade-offs \cite{pali,22bscale}. Increasing their capacity typically induces growth in compute cost and inference-time latency. This issue becomes even more acute for high-resolution images or long videos, where moderate increases in encoder cost compound into substantial end-to-end overhead.

Mixture-of-Experts (MoE) offers an alternative scaling mechanism by increasing the total model capacity while only activating a small subset of experts for any given input, thus decoupling model capacity from compute cost \cite{original_moe}. MoE has been widely used in large language models (LLMs) \cite{deepseekv3,gptoss,kimik2,qwen3,mixtral} and has carried over to VLMs \cite{qwen3vl,kimi2p5,seed1p5vl,moe_llava,gemini2p5}. In contrast, applying MoE to vision encoders, particularly to the prevailing contrastive language-image pretraining (CLIP)~\cite{clip}, remains less settled. While prior research provides several attempts on developing MoE vision encoders, none has closed the gap to State-of-the-Art (SOTA) performance on core vision benchmarks \cite{moe_cumo,clip_moe,liemoe,moe_clipup,vmoe}. 

In this work, we present a systematic study of MoE architecture design in vision encoders across multiple scales. We observe that conventional MoE \cite{original_moe}, \ie, treating MoE as a drop-in replacement for dense MLP blocks, provides only limited gains. In comparison, MoE vision encoders with fine-grained expert designs achieve larger, more consistent improvements. This aligns with insights from recent work on MoE-based LLMs \cite{deepseekv3}. MoE with fine-grained expert design is also well matched to the nature of encoding vision information, where a vision encoder is responsible for supporting a wide variety of visual concepts, styles, and domains. We further propose a variant of loss-free balancing \cite{aux_free} to balance expert utilization during training. One practical challenge is the system overhead introduced by memory-bound operations in MoE computations, which can hurt the latency benefits implied by sparsely activated parameters. We therefore develop an MoE kernel in Triton \cite{triton} to overcome this. 

\begin{figure}[htbp]
  \centering
  \includegraphics[height=5cm]{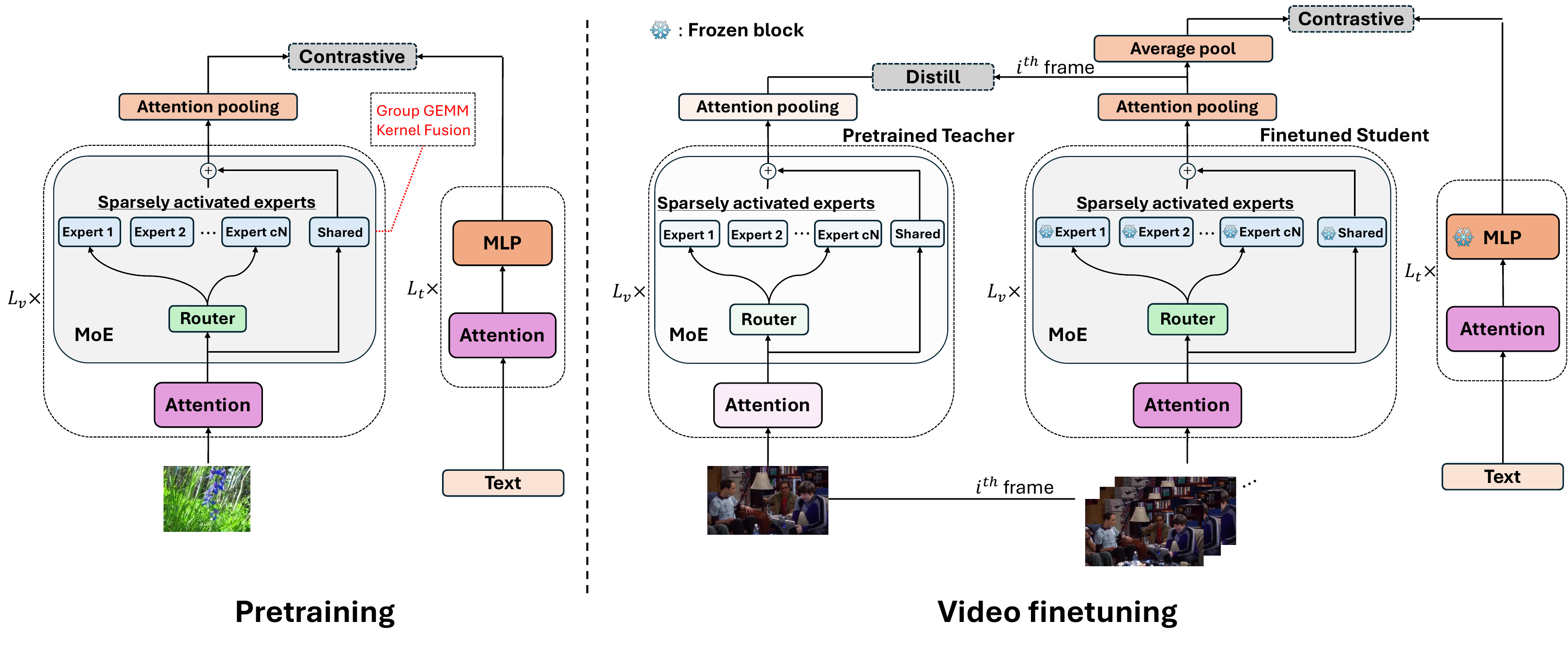}
  \caption{Illustration of MoE-ViE architecture and training pipeline. MoE-ViE adopts a fine-grained MoE design with optimized MoE kernels for latency reduction. We first perform contrastive pretraining on large-scale image-text pairs. During video finetuning, we introduce frame-level distillation and expert freezing to preserve the pretrained image understanding capabilities.}
  \label{overall}
\end{figure}

Beyond these architectural choices, we focus on refining our training strategy to emphasize building a unified image and video encoder. We first pretrain on a vast set of image-text pairs via contrastive training and then, following \cite{pe}, we finetune with video data to enhance the model's capability in video understanding. However, we observe that vanilla video finetuning can shift representations in ways that degrade our pretrained image capabilities. We mitigate this by introducing a frame-level distillation during video finetuning and by freezing MoE experts. Together, we are able to preserve the image representation learned during pretraining while allowing video supervision to strengthen video representation. Fig. \ref{overall} summarizes our MoE architecture and training pipeline.

Our investigation yields several key findings:
\begin{itemize}
\item We show that with appropriate architectural design, vision encoders with MoE consistently outperform their dense counterparts across multiple scales. This confirms the efficacy of sparse scaling for CLIP-style vision encoders.
\item We demonstrate that the MoE design proposed in this work substantially improves over prior works on MoE vision encoders across model sizes.
\item Our trained MoE-ViE achieves SOTA results at all scales on zero-shot image and video benchmarks. Our largest encoder achieves results comparable to the SOTA dense encoder PE\textsubscript{core}G, which is $1.7\times$ larger.
\item  Our MoE kernel provides $>2.5\times$ speedup in inference latency. It can be used as a drop-in implementation for different MoE topologies.
 \item Aligned with an LLM, MoE-ViE yields strong downstream results, and matches or exceeds other SOTA encoders in comparable settings.
\end{itemize}

\section{Vision Encoder with MoE}

In this section, we first review the MoE formulation, then present our studies on introducing MoE into CLIP vision encoders, including the architectural design and the control of expert utilization.

\subsection{MoE Preliminaries}

A typical MoE layer replaces the dense MLP block with $N$ experts $\{E_i\}_{i=1}^{N}$. Given an input token $x\in\mathbb{R}^{d}$, a router produces logits
\begin{equation}
s = W_r x \in \mathbb{R}^{N}
\end{equation}
followed by selecting a subset of experts via top-$k$ routing. The standard Softmax-based gating computes
\begin{equation}
\label{router_eq}
g(x) = \mathrm{Softmax}(\text{top-}k(s))
\end{equation}
Denote $\mathcal{T}(x)\subset\{1,\dots,N\}$ to be indices of selected experts, the MoE output is 
\begin{equation}
    y = \sum_{e\in \mathcal{T}(x)}{g_e(x)E_e(x)}  
\end{equation}
MoE increases the parameter count with $N$ while keeping the per-token compute proportional to $k$ experts, providing a favorable capacity vs. compute trade-off when $k\ll N$.

\subsection{MoE for Vision Encoder}

Adapting MoE to CLIP vision encoders raises three key design questions: (i) what expert topology is most appropriate for vision representations; (ii) how to ensure balanced expert utilization; (iii) how to realize the theoretical efficiency gains promised by MoE. We address these questions in the following sections.

\subsubsection{Architecture Design}

Prior CLIP MoE works replace each dense MLP with $N$ experts whose capacity matches the original MLP \cite{clip_moe,moe_clipup,liemoe,moe_cumo}. We instead adopt a fine-grained expert design, \ie, each expert has reduced hidden width relative to the original dense MLP, enabling more experts under a comparable compute budget. This choice is motivated by the heterogeneous nature of visual features (\eg, color, shape, spatial layout) and by findings from LLMs that finer expert granularity yields more efficient scaling \cite{deepseekmoe}. We apply MoE only to the vision tower and keep the text tower unchanged, since only the vision tower is used in any subsequent VLM alignment. 

In addition to specialized experts, we include $m$ shared experts that are always active, inspired by the idea of retaining a global thumbnail alongside high-resolution tiles in \cite{llava_next} as well as the success in LLMs \cite{deepseekmoe}. The shared experts provide a persistent pathway for global context, while the routed experts capture conditional, token-specific transformations. We demonstrate expert specialization in more detail in Section~\ref{moe_analysis}.

We further replace the standard MoE Softmax gating with a Sigmoid function to avoid potential expert competition \cite{sigmoid2,sigmoid3}. Given logits $s$, we compute\footnote{Some works, \eg, \cite{vmoe}, switch the order of Softmax and top-$k$ selection to enable gradient flow in top-$1$ scenario, which is not needed for Sigmoid. For consistency with recent MoE models \cite{deepseekv3,gshard}, we also adopt this switch.}
\begin{align}
    g(x) &= \text{top-}k(\mathrm{Sigmoid}(s))    \\
    g_e'(x) &= \frac{g_e(x)}{\sum_{i\in \mathcal{T}(x)} g_i(x)}
\end{align}
where $g'(x)$ renormalizes the selected scores to keep the mixture scale stable. 

Let there be $cN$ experts in total, where each expert has hidden width reduced by $c$. We reserve the first $m$ experts as shared and route over the remaining specialized experts, \ie, $\mathcal{T}(x)\subset\{m{+}1,\dots,cN\}$. The layer output is
\begin{equation}
    y = \sum_{e\in \mathcal{T}(x)}{g'_e(x)E_e(x)}+\lambda \sum_{e=1}^m{E_e(x)}
\end{equation}
where hyperparameter $\lambda$ controls the weight of shared experts. We set $\lambda=1$ in all experiments for simplicity, as our model is not sensitive to this choice. Additional details are provided in Appendix~\ref{appx_shard_experts}.

Table~\ref{different_moe} compares MoE designs trained on 12.8B samples from MetaCLIP \cite{metaclip}. We observe consistent improvements from the introduction of MoE, and the fine-grained design leads to stronger performance. 

\begin{table}[htbp]
\centering
\caption{Comparison of different MoE designs in vision encoders.}
\label{different_moe}
\fontsize{6.5}{8}\selectfont
\setlength{\tabcolsep}{2.5pt}
\newcommand{\gc}{\cellcolor{gray!10}}
\begin{tabular}{l c c c c c c c c c c c c c}
\toprule
& & & \multicolumn{6}{c}{\textbf{General Classification}} & \multicolumn{5}{c}{\textbf{Image Retrieval}} \\
\cmidrule(lr){4-9} \cmidrule(lr){10-14}
\textbf{Model}
& \rotatebox{90}{\shortstack{Total\\params}}
& \rotatebox{90}{\shortstack{Activated\\params}}
& \rotatebox{90}{Avg.\ Class.}
& \rotatebox{90}{\shortstack{{\fontsize{7}{8}\selectfont ImageNet}\\{\fontsize{3}{4}\selectfont Val \cite{imagenet1k}}}}
& \rotatebox{90}{\shortstack{{\fontsize{7}{8}\selectfont ImageNet}\\{\fontsize{3}{4}\selectfont v2 \cite{imagenetv2}}}}
& \rotatebox{90}{{\fontsize{7}{8}\selectfont ObjectNet \cite{objectnet}}}
& \rotatebox{90}{\shortstack{{\fontsize{7}{8}\selectfont ImageNet}\\{\fontsize{3}{4}\selectfont Adversarial \cite{imageneta}}}}
& \rotatebox{90}{\shortstack{{\fontsize{7}{8}\selectfont ImageNet}\\{\fontsize{3}{4}\selectfont Renditions \cite{imagenetr}}}}
& \rotatebox{90}{Avg.\ Retrieval}
& \rotatebox{90}{\shortstack{{\fontsize{7}{8}\selectfont COCO}\\{\fontsize{3}{4}\selectfont T$\rightarrow$I \cite{coco}}}}
& \rotatebox{90}{\shortstack{{\fontsize{7}{8}\selectfont COCO}\\{\fontsize{3}{4}\selectfont I$\rightarrow$T \cite{coco}}}}
& \rotatebox{90}{\shortstack{{\fontsize{7}{8}\selectfont Flickr30k}\\{\fontsize{3}{4}\selectfont T$\rightarrow$I \cite{flickr}}}}
& \rotatebox{90}{\shortstack{{\fontsize{7}{8}\selectfont Flickr30k}\\{\fontsize{3}{4}\selectfont I$\rightarrow$T \cite{flickr}}}} \\
\midrule
ViT-B/32 dense & 0.1B & 0.1B &  \gc 55.8& 67.5 & 59.3 & 50.6 & 27.8 & 73.7 & \gc 58.4& 36.3 & 54.4 & 63.3 & 79.6 \\
ViT-B/32 conventional MoE & 0.5B & 0.1B & \gc 59.9& 70.6 & 62.3 & 54.3 & 34.7 & 77.7& \gc 59.7& 37.8 & 56.3 & 64.4 & 80.2 \\
MoE-ViE-B/32 & 0.5B & 0.1B &\gc \textbf{63.2}& \textbf{72.9}	&	\textbf{64.9}& \textbf{57.7}&	\textbf{40.4} &	\textbf{80.0}	&\gc 	\textbf{61.4}&	\textbf{38.9}& \textbf{56.6}	&\textbf{66.5}	& \textbf{83.5} \\
\midrule
ViT-L/16 & 0.3B & 0.3B & \gc78.0 &79.7	&72.6 &74.9	&72.4 &90.5& \gc 68.9&  45.8 &63.8	&75.9 &90.3  \\
ViT-L/16 conventional MoE & 1.7B & 0.3B & \gc 78.5	&80.2 &72.9	&73.5 &74.9	&90.8& \gc 69.0 &45.5	&63.4 &76.1	&90.8 \\
MoE-ViE-L/16 & 1.7B & 0.3B & \gc \textbf{79.6} & \textbf{80.7}	&\textbf{74.1} & \textbf{76.5}	& \textbf{75.4} & \textbf{91.5}& \gc \textbf{69.3} & \textbf{46.0}	& \textbf{63.8} & \textbf{76.3}	& \textbf{91.0} \\
\bottomrule
\end{tabular}
\end{table}

\subsubsection{Load Balancing}
\label{aux_design}
Compared to prior works designing various auxiliary losses \cite{liemoe,vmoe}, optimized jointly with the main training objective, we propose a variant of the loss-free approach \cite{aux_free}, which decouples the control of expert utilization from the training objective. In loss-free balancing, an extra bias term $b$ is introduced to the router. Given routing logits $s$, the router now selects experts by
\begin{equation}
    g(x) = \text{top-}k(\mathrm{Sigmoid}(s)+b) 
\end{equation}
During training, the bias for expert $e$ is updated using the observed token load. Let $t_e$ be the number of tokens routed to expert $e$ in the current iteration, and let $\mu_t$ denote the mean token count across experts. The update rule is
\begin{equation}
    b_e = b_e - \alpha\ \mathrm{sign}(t_e-\mu_t)
\end{equation}
with step size $\alpha$. We notice, however, that the $\mathrm{sign}(\cdot)$ update applies a constant magnitude correction (\eg, $1$ or $-1$) regardless of deviation, which can cause oscillations when routing is close to balanced. Therefore, we propose to replace $\mathrm{sign}(\cdot)$ with z-score, which scales the correction by the degree of imbalance and keeps it normalized for stability. The update of the bias term then becomes
\begin{equation}
    b_e = b_e - \alpha \frac{t_e-\mu_t}{\sigma_t}
\end{equation}
where $\sigma_t$ is the standard deviation of $\{t_e\}$. We present ablations on different load balancing strategies in Section~\ref{ablation}.

\subsubsection{Scalability}

\begin{figure}[htbp]
  \centering
  \includegraphics[height=5cm]{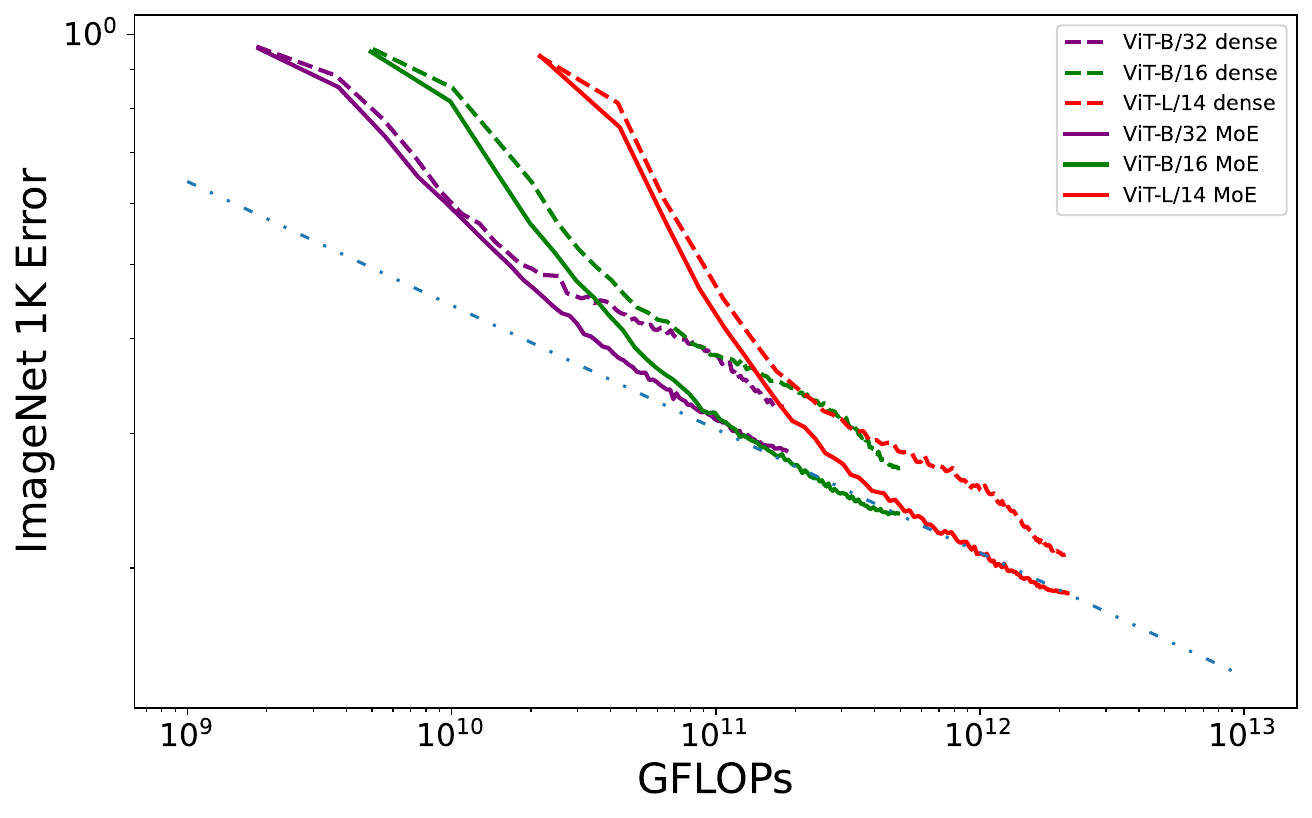}
  \caption{MoE-ViE vs. dense models at multiple scales. MoE-ViE consistently outperforms their dense counterparts under the same compute budget.}
  \label{scaling_law}
\end{figure}

We then explore the scalability of MoE-ViE under contrastive pretraining. We train our models at multiple scales. At each scale, we train both the baseline dense model and MoE-ViE. All models are trained for $12.8$B samples on MetaCLIP \cite{metaclip} at $224\times224$ resolution, and evaluated using ImageNet-1K classification accuracy \cite{imagenet1k}. Fig. \ref{scaling_law} summarizes the results. Across all scales, MoE-ViE consistently outperforms its dense counterpart under the same compute budget, indicating a desired model capacity vs. compute trade-off. This motivates us to further scale up MoE-ViE as detailed in Section~\ref{main_results}.

\subsection{Latency Optimization}
\label{optimization}

\subsubsection{Inherent MoE Efficiency} Unlike dense networks where FLOPs scale linearly with total parameter count, MoE maintains constant active FLOPs per token by routing inputs to a small subset of experts. Thus, MoE theoretically offers superior representation (via more parameters) without an inherent latency penalty, as its compute cost is governed only by the number of active parameters.

\subsubsection{Optimized Kernel} 
While MoE is theoretically compute-efficient, naive implementations (\eg, sequentially looping over experts in PyTorch) often under-utilize GPUs, mainly due to three factors: (1) fragmenting compute into many small general matrix multiply (GEMM) operations reduces arithmetic intensity; (2) dynamic routing introduces CPU–GPU synchronization to determine per-expert workloads, creating idle hardware “bubbles”; (3) larger intermediate tensors amplify HBM traffic, making memory bandwidth the latency bottleneck.

To realize MoE’s theoretical efficiency in practice, we implement a specialized Triton \cite{triton} kernel with two key optimizations: First, \textit{Grouped GEMM} aggregates all expert matrix multiplications (MatMuls) into a single, unified GPU task, enabling parallel processing across experts and eliminating repeated CPU–GPU handshakes. Second, \textit{Kernel Fusion} combines MatMuls and non-linear activations (\eg, GeLU) into one pass, minimizing round-trips to HBM and reducing data movement. More details are included in Appendix~\ref{appx_kernel}.

These optimizations are not auxiliary architectural tweaks. Instead, they are the necessary implementation mechanisms to faithfully translate MoE’s algorithmic efficiency into practical, hardware-efficient execution.

\section{Robust Video Finetuning}
\label{video_ft}

Following \cite{pe}, we first pretrain the model on image-text pairs and then finetune on video-text pairs to build a unified vision encoder for both image and video understanding. However, in our experiments, naively finetuning on video data substantially degrades image performance, indicating severe forgetting. Mixing image data into the video stage partially recovers image accuracy, but we find it also limits gains on video understanding (Section~\ref{ablation}).

Recognizing that most VLMs use frame-based vision encoders to encode videos frame by frame \cite{plm,omnivinci}, we regularize the video finetuning with frame-level distillation. We keep a copy of the image-pretrained model as a teacher. For a given video, we randomly sample a frame and feed it through the student and teacher vision towers to obtain logits $S$ and $T$, and minimize the cosine distance 
\begin{equation}
\label{similarity}
    \mathcal{L}_d = 1 - \cos(S, T)
\end{equation}
The final loss becomes
\begin{equation}
    \mathcal{L}=\mathcal{L}_c+\beta\mathcal{L}_d
\end{equation}
where $\mathcal{L}_c$ denotes contrastive training loss, and $\beta$ is the weight of distillation.

We further reduce forgetting by freezing the MoE experts during video finetuning. This aligns with findings that MLP layers store models' acquired knowledge, while attention layers are primarily responsible for feature interaction \cite{attn_insights}. Since our vision encoder operates on individual frames, we do not expect video finetuning to require large changes to the per-frame visual vocabulary. Instead, the adaptation should mainly update how frame features are aggregated for video-specific alignment. 

One issue is that $\mathcal{L}_c$ updates both towers. So when finetuning on videos, the text tower may drift away from the alignment established during pretraining, while $\mathcal{L}_d$ encourages the vision tower to stay close to its image-pretrained teacher, creating an optimization mismatch. We find that an efficient way to eliminate this is to freeze the MLP layers in the text tower as well. Our intuition is that the text tower has already trained on substantially larger and more diverse data, and our primary goal in this stage is to adapt the vision encoder to videos. 

\section{Experiments}

We develop a series of MoE-ViE at various scales by building on top of the standard CLIP vision encoder architecture, where we replace the MLP modules in the vision tower with our proposed MoE blocks. We exclude the first transformer block from this replacement, as this first layer captures general, low-level, visual information, and we did not observe any gains by specializing it with MoE in our experiments. 

\subsection{Zero-Shot Benchmarks}
\label{main_results}

\subsubsection{Setup} We conduct contrastive pretraining for MoE-ViE on $3.5$B image-text pairs, comprised of $2$B data from MetaCLIP \cite{metaclip}, and $1.5$B proprietary data. Unless otherwise specified, all of our MoE variants use $32$ experts for each MoE layer to maintain a practical compute budget. Each expert is designed to be $1/4$ of the corresponding dense MLP. We include more studies on the impact of the granularity in Appendix~\ref{appx_moe_scaling} and on the number of experts in Appendix~\ref{moe_scaling}.

We first train MoE-ViE-B and MoE-ViE-L models with $1/8$ of experts activated, and then train an MoE-ViE-H with $1/4$ experts activated to explore further scaling. Following \cite{pe}, we perform progressive resolution training, \ie, we increase image resolution as training progresses, for better training efficiency. After pretraining, we perform a short warmup using the pretraining data augmented with CC12M \cite{cc12m} and TreeOfLife \cite{treeoflife}. For video finetuning, we uniformly sample 8 frames per video and use the curated video-text data from \cite{pe,merlot}. Full training and evaluation details are provided in Appendices~\ref{appx_train_spec} and~\ref{appx_eval_spec}, respectively. Additional linear probing and LLM alignment results are reported in Appendix~\ref{appx_additional_results}.

\subsubsection{Video Benchmarks}

We report the zero-shot video understanding capability of MoE-ViE in Table~\ref{general_video} on both classification and retrieval tasks. MoE-ViE achieves SOTA results at all evaluated scales, and even outperforms the larger PE\textsubscript{core}G model, which indicates that MoE scaling transfers effectively from image pretraining to our proposed video finetuning.

\begin{table}[htbp]
\centering
\caption{Comparison of zero-shot video benchmarks between MoE-ViE and other previous SOTA vision encoders in the literature.}
\label{general_video}
\fontsize{6.5}{8}\selectfont
\newcommand{\gc}{\cellcolor{gray!15}}
\begin{tabular}{l c c c c c c c c c c c}
\toprule
& & & & \multicolumn{5}{c}{\textbf{Video Classification}} & \multicolumn{3}{c}{\textbf{Video Retrieval}} \\
\cmidrule(lr){5-9} \cmidrule(lr){10-12}
\textbf{Model}
& \rotatebox{90}{\shortstack{Total\\params}}
& \rotatebox{90}{\shortstack{Activated\\params}}
& \rotatebox{90}{Resolution}
& \rotatebox{90}{Avg.\ Class.}
& \rotatebox{90}{K400 \cite{kinetics}}
& \rotatebox{90}{K600 \cite{kinetics}}
& \rotatebox{90}{UCF101 \cite{ucf}}
& \rotatebox{90}{HMDB \cite{hmdb}}
& \rotatebox{90}{Avg.\ Retrieval}
& \rotatebox{90}{\shortstack{{\fontsize{7}{8}\selectfont MSR-VTT}\\{\fontsize{3}{4}\selectfont T$\rightarrow$V \cite{msr_vtt}}}}
& \rotatebox{90}{\shortstack{{\fontsize{7}{8}\selectfont MSR-VTT}\\{\fontsize{3}{4}\selectfont V$\rightarrow$T \cite{msr_vtt}}}} \\
\midrule
SigLIP2-B/16 \cite{siglip2} & 0.1B & 0.1B & $224$ & \gc 59.5& 58.7 & 55.0 & 82.0 & 42.3 & \gc 34.3& 38.5 & 30.1 \\
PE\textsubscript{core}B/16 \cite{pe} & 0.1B & 0.1B & $224$ & \gc 65.9& 65.6 & 65.1 & \textbf{84.6} & 48.2 & \gc 47.5& 47.6 & \textbf{47.3} \\
MoE-ViE-B/16 & 0.5B & 0.1B & $224$ & \gc \textbf{67.5} & \textbf{68.3} & \textbf{65.6} & 84.5 & \textbf{51.5} & \gc \textbf{47.6}& \textbf{47.9}& 47.2  \\
\midrule
SigLIP2-L/16 \cite{siglip2} & 0.3B & 0.3B & $384$ & \gc 66.0& 65.3 & 62.5 & 86.7 & 49.3 & \gc 36.5& 41.5  & 31.4 \\
PE\textsubscript{core}L/14 \cite{pe} & 0.3B & 0.3B & $336$ & \gc 72.9& 73.4 & \textbf{72.7} & 87.1 & \textbf{58.5} & \gc \textbf{50.2}& 50.3 & \textbf{50.1} \\
MoE-ViE-L/16 & 1.7B & 0.3B & $384$ & \gc \textbf{73.4}& \textbf{74.5} & 71.8 & \textbf{89.2} & 57.9 & \gc \textbf{50.2}& \textbf{50.5} & 49.8 \\
\midrule
InternVL-C/16 \cite{internvl} & 5.5B & 5.5B & $224$ & \gc -& 69.1 & 68.9 & - & - & \gc 42.5& 44.7 & 40.2 \\
SigLIP2-g-opt/16 \cite{siglip2} & 1.1B & 1.1B & $384$ & \gc 69.8& 69.8 & 67.0 & 90.7 & 51.8 & \gc 38.7& 43.1 & 34.2 \\
PE\textsubscript{core}G/14 \cite{pe} & 1.9B & 1.9B & $448$ & \gc 76.0& \textbf{76.9} & \textbf{75.1} & 90.7 & 61.1 & \gc \textbf{50.6}& 51.2 & \textbf{49.9} \\
MoE-ViE-H/14 & 3.5B & 1.1B & $448$ & \gc \textbf{76.5} & \textbf{76.9} & 75.1 & \textbf{92.8} & \textbf{61.3} & \gc \textbf{50.6} & \textbf{51.6} & 49.5 \\
\bottomrule
\end{tabular}
\end{table}

\subsubsection{General Image Benchmarks}

Table~\ref{general_image} summarizes zero-shot performance on image classification and retrieval benchmarks. We compare MoE-ViE to both SOTA dense vision encoders and to prior CLIP MoE vision encoders. MoE-ViE achieves the best results at each scale among models with comparable sizes. In particular, MoE-ViE demonstrates consistently high accuracy on the challenging ImageNet-A \cite{imageneta} benchmark, with $+5.6\%$ over PE\textsubscript{core}B, $+0.2\%$ over PE\textsubscript{core}L, and even $+0.6\%$ over PE\textsubscript{core}G which has $1.7\times$ as many activated parameters as MoE-ViE-H.

\begin{table}[htbp]
\centering
\caption{Comparison of zero-shot image classification and retrieval between MoE-ViE and other SOTA vision encoders in the literature.}
\label{general_image}
\fontsize{6.5}{8}\selectfont
\setlength{\tabcolsep}{2.5pt}
\newcommand{\gc}{\cellcolor{gray!10}}
\resizebox{\textwidth}{!}{
\begin{tabular}{l c c c c c c c c c c c c c c}
\toprule
& & & & \multicolumn{6}{c}{\textbf{General Image Classification}} & \multicolumn{5}{c}{\textbf{Image Retrieval}} \\
\cmidrule(lr){5-10} \cmidrule(lr){11-15}
\textbf{Model}
& \rotatebox{90}{\shortstack{Total\\params}}
& \rotatebox{90}{\shortstack{Activated\\params}}
& \rotatebox{90}{Resolution}
& \rotatebox{90}{Avg.\ Class.}
& \rotatebox{90}{\shortstack{{\fontsize{7}{8}\selectfont ImageNet}\\{\fontsize{3}{4}\selectfont Val \cite{imagenet1k}}}}
& \rotatebox{90}{\shortstack{{\fontsize{7}{8}\selectfont ImageNet}\\{\fontsize{3}{4}\selectfont v2 \cite{imagenetv2}}}}
& \rotatebox{90}{{\fontsize{7}{8}\selectfont ObjectNet \cite{objectnet}}}
& \rotatebox{90}{\shortstack{{\fontsize{7}{8}\selectfont ImageNet}\\{\fontsize{3}{4}\selectfont Adversarial \cite{imageneta}}}}
& \rotatebox{90}{\shortstack{{\fontsize{7}{8}\selectfont ImageNet}\\{\fontsize{3}{4}\selectfont Renditions \cite{imagenetr}}}}
& \rotatebox{90}{Avg.\ Retrieval}
& \rotatebox{90}{\shortstack{{\fontsize{7}{8}\selectfont COCO}\\{\fontsize{3}{4}\selectfont T$\rightarrow$I \cite{coco}}}}
& \rotatebox{90}{\shortstack{{\fontsize{7}{8}\selectfont COCO}\\{\fontsize{3}{4}\selectfont I$\rightarrow$T \cite{coco}}}}
& \rotatebox{90}{\shortstack{{\fontsize{7}{8}\selectfont Flickr30k}\\{\fontsize{3}{4}\selectfont T$\rightarrow$I \cite{flickr}}}}
& \rotatebox{90}{\shortstack{{\fontsize{7}{8}\selectfont Flickr30k}\\{\fontsize{3}{4}\selectfont I$\rightarrow$T \cite{flickr}}}} \\
\midrule
LIMOE-B/16 \cite{liemoe} & 0.5B & 0.1B & $224$ & \gc -& 73.7 & - & - & - & - & \gc -& 36.2 & 51.3 &  -& - \\
CLIP-UP-B/16\cite{moe_clipup} & 0.5B & 0.2B & $224$ & \gc -& 76.9 & - & - & - & - & \gc 74.2 & \textbf{52.1} & \textbf{71.5} & \textbf{80.9} & 92.3 \\
SigLIP2-B/16 \cite{siglip2} & 0.1B & 0.1B & $224$ & \gc 74.0& 78.2& 71.4 & 73.6 & 55.0 & \textbf{91.7} & \gc 73.7& \textbf{52.1} & 68.9 & 80.7 & 93.0 \\
PE\textsubscript{core}B/16 \cite{pe} & 0.1B & 0.1B & $224$ & \gc 74.6&	78.4	&71.7	&71.9	&62.4	&88.7 & \gc 74.3& 50.9 & 71.0 & 80.8 & \textbf{94.4} \\
MoE-ViE-B/16 & 0.5B & 0.1B & $224$ & \gc \textbf{76.8}	& \textbf{79.3}&	\textbf{72.5}	& \textbf{74.4}	& \textbf{68.0}	& 89.9 &\gc \textbf{74.4}&	\textbf{52.1}&	70.7	&\textbf{80.9}	&93.9  \\
\midrule
CLIP-MoE-L/14 \cite{clip_moe} & 0.9B & 0.5B & $336$ & \gc -& 74.6 & 68.5 & 33.5 & - & - & \gc 53.6& 46.8 & 65.0 & 42.1 & 60.5 \\
LIMOE-L/16 \cite{liemoe} & 1.7B & 0.3B & $224$ & \gc -& 78.6 & - & - & - & - & \gc -& 39.6 & 55.7 & - & - \\
CLIP-UP-L/14 \cite{moe_clipup} & 1.7B & 0.5B & $224$ & \gc -& 81.2 & - & - & - & - & \gc 75.5& 53.9 & 73.8 & 82.0 & 92.4 \\
SigLIP2-L/16 \cite{siglip2} & 0.3B & 0.3B & $384$ & \gc 85.0& 83.1 & 77.4 & 84.4 & 84.3 & \textbf{95.7} & \gc76.7 & 55.3 & 71.4 & 85.0 & 95.2 \\
PE\textsubscript{core}L/14 \cite{pe} & 0.3B & 0.3B & $336$ & \gc 86.1& 83.5 & \textbf{77.9} & 84.7 & \textbf{89.0} & 95.2 & \gc \textbf{78.8}& 57.1 & \textbf{75.9} & 85.5 & \textbf{96.6} \\
MoE-ViE-L/16 & 1.7B & 0.3B & $384$ & \gc \textbf{86.2}& \textbf{83.6} & \textbf{77.9} & \textbf{85.0}& \textbf{89.0} & 95.4 & \gc \textbf{78.8}& \textbf{57.2}	& \textbf{75.9}	&\textbf{85.7}	&96.5  \\
\midrule
EVA 18B/14 \cite{eva-clip-18b} & 17.5B & 17.5B & $224$ & \gc 85.4& 83.8 & 77.9 & 82.2 & 87.3 & 95.7 & \gc 77.5& 56.2 & 73.6 & 83.3 & 96.7 \\
InternVL-C/16 \cite{internvl} & 5.5B & 5.5B & $224$ & \gc 84.1& 83.2 & 77.3 & 80.6 & 83.8 & 95.7 & \gc 78.6& 58.6 & 74.9 & 85.0 & 95.7 \\
SigLIP2-g-opt/16 \cite{siglip2} & 1.1B & 1.1B & $384$ & \gc 88.0 & 85.0 & 79.8 & 88.0 & 90.5 & 96.6 & \gc 77.6 & 56.1 & 72.8 & \textbf{86.0} & 95.4 \\
PE\textsubscript{core}G/14 \cite{pe} & 1.9B & 1.9B & $448$ & \gc \textbf{88.6} & \textbf{85.4} & \textbf{80.2} & \textbf{88.2} & 92.6 & \textbf{96.5} & \gc \textbf{78.9} & \textbf{58.1} & \textbf{75.4} & 85.7 & \textbf{96.2} \\
MoE-ViE-H/14 & 3.5B & 1.1B & $448$ & \gc 88.3 & 85.1 & 80.0 & 87.0 & \textbf{93.2} & 96.2 & \gc 78.2 & 56.8 & 74.6 & 85.4 & 96.0 \\
\bottomrule
\end{tabular}
}
\end{table}

\subsubsection{Fine-grained Image Benchmarks}

Table~\ref{finegrain_image} presents the results for image classification on fine-grained categories. We see MoE-ViE excels in these tasks. Not only do our MoE-ViE models achieve SOTA results compared with similarly sized models, here, again, our largest model MoE-ViE-H outperforms PE\textsubscript{core}G, whose activated size is $1.7\times$ larger. Additionally, our model demonstrates high accuracy on OCR benchmarks, as evaluated on TextCaps \cite{textcap}. We attribute these achievements to the nature of MoE, applied with the correct topology, enabling experts to specialize in different tasks.

\begin{table}[htbp]
\centering
\caption{Comparison of zero-shot fine-grained image classification and OCR benchmarks between MoE-ViE and SOTA dense vision encoders. We re-evaluated part of the OCR results for SigLIP2 and PE if not reported in \cite{siglip2} and \cite{pe}.}
\label{finegrain_image}
\fontsize{6.5}{8}\selectfont
\newcommand{\gc}{\cellcolor{gray!15}}
\begin{tabular}{l c c c c c c c c c c c c}
\toprule
& & & & \multicolumn{6}{c}{\makebox[0pt]{\textbf{Fine-grained Img Classification}}}
 & \multicolumn{3}{c}{\textbf{OCR}} \\
\cmidrule(lr){5-10} \cmidrule(lr){11-13}
\textbf{Model}
& \rotatebox{90}{\shortstack{Total\\params}}
& \rotatebox{90}{\shortstack{Activated\\params}}
& \rotatebox{90}{Resolution}
& \rotatebox{90}{Avg.\ Class.}
& \rotatebox{90}{Food101 \cite{food}}
& \rotatebox{90}{Flowers \cite{flowers}}
& \rotatebox{90}{Country211 \cite{country}}
& \rotatebox{90}{Aircrafts \cite{aircraft}}
& \rotatebox{90}{Cars \cite{cars}}
& \rotatebox{90}{Average OCR}
& \rotatebox{90}{\shortstack{{\fontsize{7}{8}\selectfont Text Cap}\\{\fontsize{3}{4}\selectfont T$\rightarrow$I \cite{textcap}}}}
& \rotatebox{90}{\shortstack{{\fontsize{7}{8}\selectfont Text Cap}\\{\fontsize{3}{4}\selectfont I$\rightarrow$T \cite{textcap}}}} \\
\midrule
SigLIP2-B/16 \cite{siglip2} & 0.1B & 0.1B & $224$ & \gc 69.2& 92.8 & 85.7 & 19.2 & 54.8 & \textbf{93.4} & \gc 71.5& 72.3 & 70.7 \\
PE\textsubscript{core}B/16 \cite{pe} & 0.1B & 0.1B & $224$ & \gc 71.7& 92.5 & 86.5 & 30.5 & 57.0 & 92.1 & \gc 71.6&  72.3&70.9  \\
MoE-ViE-B/16 & 0.5B & 0.1B & $224$ & \gc \textbf{73.5}	&\textbf{94.2}	&\textbf{87.3}	&\textbf{34.9}	&\textbf{58.2}&	93.1 & \gc \textbf{72.5}& \textbf{72.7} & \textbf{72.3} \\
\midrule
SigLIP2-L/16 \cite{siglip2} & 0.3B & 0.3B & $384$ & \gc 76.1&  96.1&  \textbf{90.0}& 31.6 & 67.0 & \textbf{95.8} & \gc 79.2& \textbf{80.2} & 78.2\\
PE\textsubscript{core}L/14 \cite{pe} & 0.3B & 0.3B & $336$ & \gc 78.1& 96.2 & 87.2 & 45.6 & \textbf{67.8} & 93.7 & \gc 79.2& 79.8 & 78.5 \\
MoE-ViE-L/16 & 1.7B & 0.3B & $384$ & \gc \textbf{78.9}& \textbf{96.7} & 89.2 & \textbf{47.8}& 65.7 & 94.9 & \gc \textbf{79.7}& 80.0 & \textbf{79.3} \\
\midrule
EVA 18B/14 \cite{eva-clip-18b} & 17.5B & 17.5B & $224$ & \gc 75.9& 95.8 & 86.0 & 43.1 & 59.7 & 94.9 & \gc- &  -&  -\\
InternVL-C/16 \cite{internvl} & 5.5B & 5.5B & $224$ & \gc 72.8& 95.3 & 85.8 & 35.1 & 53.3 & 94.4 & \gc -& - & 72.3 \\
SigLIP2-g-opt/16 \cite{siglip2} & 1.1B & 1.1B & $384$ & \gc 79.6 & 97.0 & 91.5 & 40.1 & 73.6 & \textbf{95.9} & \gc 79.8& 80.3 & 79.2 \\
PE\textsubscript{core}G/14 \cite{pe} & 1.9B & 1.9B & $448$ & \gc 83.8 & 96.9 & 91.4 & \textbf{57.6} & 78.2 & 94.7 & \gc 79.1& 79.3 & 78.8 \\
MoE-ViE-H/14 & 3.5B & 1.1B & $448$ & \gc \textbf{83.9} & \textbf{97.1} & \textbf{92.4} & 54.3 & \textbf{80.1} & 95.5 & \gc \textbf{80.4} & \textbf{80.6} & \textbf{80.2} \\
\bottomrule
\end{tabular}
\end{table}

\subsection{VLM Alignment}
\label{subsec:align}
Since VLMs represent the primary downstream application of vision encoders, evaluating alignment quality is essential to verify that improvements in visual representations translate to gains on end-user tasks such as visual question answering, video understanding, and captioning. Our alignment training follows a standard three-stage pipeline. In Stage~1 (warmup, 8k steps), we train only the two-layer MLP projectors while keeping all other parameters frozen. In Stage~2 (pre-alignment, 15k steps), we explore various training strategies, and find that freezing the encoder parameters for the first 1k iterations to stabilize the loss before transitioning to end-to-end training of all parameters yields the best downstream performance. In Stage~3 (supervised finetuning, 8k steps), we train all parameters of the model. Results are reported in Table~\ref{tab:model_performance} and Table~\ref{tab:qwen} with Llama 3.1 Instruct 8B and Qwen 2.5 VL 7B, respectively. We use the tile-based approach \cite{internvl3,molmo2} to support dynamic resolutions. 

\begin{table}[htbp]
\centering
\caption{Alignment comparison across models. Columns are grouped by task type. Llama 3.1 Instruct 8B \cite{llama3} is used as the base LLM, and we use a maximum of 4 tiles.}
\label{tab:model_performance}
\fontsize{6.5}{8}\selectfont
\setlength{\tabcolsep}{3pt}
\newcommand{\gc}{\cellcolor{gray!15}}
\resizebox{\textwidth}{!}{%
\begin{tabular}{l c c c c c c c c c c c c c c}
\toprule
& & \multicolumn{6}{c}{\textbf{Image}} & \multicolumn{4}{c}{\textbf{Video}} & \multicolumn{3}{c}{\textbf{Captioning}} \\
\cmidrule(lr){3-8} \cmidrule(lr){9-12} \cmidrule(lr){13-15}
\textbf{Model}
& \rotatebox{90}{\shortstack{Activated\\params}}
& \rotatebox{90}{Avg.\ Images}
& \rotatebox{90}{AI2D\cite{ai2d}}
& \rotatebox{90}{TextVQA\cite{textvqa}}
& \rotatebox{90}{ChartQA\cite{chartqa}}
& \rotatebox{90}{DocVQA\cite{docvqa}}
& \rotatebox{90}{Info.\ QA\cite{infovqa}}
& \rotatebox{90}{Avg.\ Video}
& \rotatebox{90}{VideoMME\cite{videomme}}
& \rotatebox{90}{MVBench\cite{mvbench}}
& \rotatebox{90}{EgoSchema\cite{egoschema}}
& \rotatebox{90}{Avg.\ Cap.}
& \rotatebox{90}{COCO\cite{coco}}
& \rotatebox{90}{NoCaps\cite{nocaps}} \\
\midrule
MetaCLIP-G/14 \cite{metaclip}         & 1.8B & \gc 61.3 & 72.8 & 65.4 & 68.1 & 61.3 & 39.1 & \gc 45.4 & 46.5 & 44.7 & 45.0 & \gc 125.5 & 134.4 & 116.5 \\
SigLIP2-g-opt/16 \cite{siglip2}       & 1.1B & \gc 59.0 & 72.4 & 70.3 & 63.1 & 55.3 & 34.0 & \gc 49.5 & 46.2 & 48.5 & 53.8 & \gc 129.1 & 137.8 & 120.3 \\
PE\textsubscript{core}G/14 \cite{pe}  & 1.9B & \gc 69.2 & 69.7 & 74.3 & 73.4 & 81.2 & 47.6 & \gc 48.6 & 46.0 & 48.7 & 51.2 & \gc 123.7 & 134.5 & 112.9 \\
InternViT2.5/14 \cite{internvl2p5}    & 5.5B & \gc 68.1 & 72.9 & 71.3 & \textbf{74.6} & 74.3 & 47.6 & \gc 48.7 & 46.0 & 49.6 & 50.6 & \gc 123.0 & 132.5 & 113.5 \\
AIMv2 3B/14 \cite{aimv2}              & 2.7B & \gc 69.8 & 72.2 & \textbf{79.2} & 73.0 & 78.2 & 46.5 & \gc 49.7 & 49.6 & 49.9 & 49.6 & \gc \textbf{130.6} & \textbf{139.7} & \textbf{121.5} \\
\midrule
MoE-ViE-H/14                          & 1.1B & \gc \textbf{75.2} & \textbf{84.9} & 75.7 & 74.2 & \textbf{86.1} & \textbf{55.0} & \gc \textbf{58.9} & \textbf{52.0} & \textbf{63.6} & \textbf{61.2} & \gc 127.5 & 135.4 & 119.6 \\
\bottomrule
\end{tabular}
}
\end{table}

\begin{table}[htbp]
\centering
\caption{Alignment comparison across models. Columns are grouped by task type. Qwen 2.5 VL 7B is used as the base LLM, and we use a maximum of 4 tiles.}
\label{tab:qwen}
\fontsize{6.5}{8}\selectfont
\setlength{\tabcolsep}{3pt}
\newcommand{\gc}{\cellcolor{gray!15}}
\resizebox{\textwidth}{!}{%
\begin{tabular}{l c c c c c c c c c c c c c c}
\toprule
& & \multicolumn{6}{c}{\textbf{Image}} & \multicolumn{4}{c}{\textbf{Video}} & \multicolumn{3}{c}{\textbf{Captioning}} \\
\cmidrule(lr){3-8} \cmidrule(lr){9-12} \cmidrule(lr){13-15}
\textbf{Model}
& \rotatebox{90}{\shortstack{Activated\\params}}
& \rotatebox{90}{Avg.\ Images}
& \rotatebox{90}{AI2D\cite{ai2d}}
& \rotatebox{90}{TextVQA\cite{textvqa}}
& \rotatebox{90}{ChartQA\cite{chartqa}}
& \rotatebox{90}{DocVQA\cite{docvqa}}
& \rotatebox{90}{Info.\ QA\cite{infovqa}}
& \rotatebox{90}{Avg.\ Video}
& \rotatebox{90}{VideoMME\cite{videomme}}
& \rotatebox{90}{MVBench\cite{mvbench}}
& \rotatebox{90}{EgoSchema\cite{egoschema}}
& \rotatebox{90}{Avg.\ Cap.}
& \rotatebox{90}{COCO\cite{coco}}
& \rotatebox{90}{NoCaps\cite{nocaps}} \\
\midrule
SigLIP2-g-opt/16 \cite{siglip2}       & 1.1B & \gc 62.7 & 75.2 & 70.3 & 71.0 & 60.4 & 36.7 & \gc 55.6 & 52.0 & 52.8 & 62.0 & \gc 130.1 & 139.0 & 121.1 \\
PE\textsubscript{core}G/14 \cite{pe}  & 1.9B & \gc 65.4 & 72.9 & 67.9 & 75.9 & 68.8 & 41.6 & \gc 54.1 & 48.7 & 52.9 & 60.8 & \gc 123.8 & 135.2 & 112.3 \\
InternViT2.5/14 \cite{internvl2p5}    & 5.5B & \gc 65.4 & 73.6 & 70.1 & \textbf{78.2} & 65.3 & 39.6 & \gc 52.7 & 50.3 &  51.1 & 56.6 & \gc 128.5 & 138.4 & 118.6 \\
AIMv2 3B/14 \cite{aimv2}              & 2.7B & \gc 67.6 & 75.2 & \textbf{74.2} & 76.7 & 70.5 &  41.4 & \gc 52.7 & 45.9 & 51.4 & 60.8 & \gc \textbf{130.6} & \textbf{139.2} & \textbf{122.0} \\
\midrule
MoE-ViE-H/14                          & 1.1B & \gc \textbf{74.0} & \textbf{86.5} & 72.3 & 73.0 & \textbf{84.1} & \textbf{54.0} & \gc \textbf{59.2} & \textbf{53.4} & \textbf{61.5} & \textbf{62.8} & \gc 129.7 & 138.6 & 120.8 \\
\bottomrule
\end{tabular}
}
\end{table}

As shown in Table~\ref{tab:model_performance} and Table~\ref{tab:qwen}, MoE-ViE-H achieves strong performance across all evaluation axes. On both image and video benchmarks, our model outperforms all baselines within the same parameter class and even surpasses SOTA models with several times more activated parameters. In captioning, although we do not achieve the highest scores, our results remain highly competitive: among models with fewer than 2B activated parameters, MoE-ViE-H is second only to SigLIP2-g-opt. We provide more analysis on alignment in Appendix~\ref{appx_alignment_results}.

\subsection{Latency Comparison}
With our kernel optimization presented in Section~\ref{optimization}, we achieve inference latency comparable to a dense model (SigLIP2-g-opt) with a similar number of active parameters, despite possessing a significantly larger pool of trainable parameters. As shown in Table~\ref{tab:performance_results}, the MoE kernel provides $>2.5\times$ speedup across different batch sizes compared to the vanilla implementation. Our MoE-ViE runs at roughly 76\% of latency compared to PE\textsubscript{core}G/14, the $1.7\times$ larger SOTA dense vision encoder, while demonstrating comparable performance on accuracy. This confirms that the representational gains of our proposed MoE-ViE are achieved without any unexpected latency regressions. Additionally, we further demonstrate that such efficiency gain is naturally translated to downstream VLMs, detailed in Appendix~\ref{appx_e2e_latency}.

\begin{table}[h]
\centering
\caption{Performance comparison between MoE-ViE and dense vision encoders, across different inference batch sizes (\textit{bsz}). Due to difference in patch size, we report latency based on the same number of input tokens (\# tokens = 576) for a fair comparison. } 
\label{tab:performance_results}
\fontsize{6.5}{8}\selectfont
\begin{tabular}{l c c c c c c}
\toprule
\textbf{} & \multirow{2}{*}{\parbox{1.5cm}{\centering \textbf{Activated\\Parameters}}} & \multirow{2}{*}{\parbox{1.5cm}{\centering \textbf{Total\\Parameters}}} & \multicolumn{4}{c}{\textbf{Latency (ms)}} \\
\cmidrule(lr){4-7}
\textbf{Model} & & & \textbf{\textit{bsz}=16} & \textbf{32} & \textbf{64} & \textbf{128} \\
\midrule
SigLIP2-g-opt/16 \cite{siglip2} & 1.1B & 1.1B & 76.39 & 141.78 & 275.94 & 549.62  \\
PE\textsubscript{core}G/14 \cite{pe} & 1.9B & 1.9B & 101.21 & 188.76 & 366.23 & 708.62 \\
\midrule
MoE-ViE-H/14 (Vanilla Implementation) & 1.1B & 3.5B & 318.76  & 448.93 & 740.84 & 1306.81  \\
MoE-ViE-H/14 (Optimized Kernel) & 1.1B & 3.5B & 82.59  & 145.82 & 276.87 & 544.96  \\
\bottomrule
\end{tabular}
\end{table}

\subsection{Ablation Study}
\label{ablation}

\subsubsection{Auxiliary Loss} Table~\ref{aux_compare} compares different expert balancing algorithms. We conduct these experiments with MoE-ViE-B with patch size $32$, which is faster to train, but exhibits the same routing dynamics as other MoE-ViE variants. As seen, the loss-free approaches (\ie, \cite{aux_free} and ours) outperform other forms on all benchmarks. This aligns with our expectation, since expert utilization is achieved without directly perturbing optimization under the contrastive training objective. Our proposed magnitude-aware method further improves over \cite{aux_free}, indicating a more stable and effective balancing control.

\begin{table}[htbp]
\centering
\caption{Comparison of different auxiliary losses for training CLIP vision encoders.}
\label{aux_compare}
\fontsize{6.5}{8}\selectfont
\newcommand{\gc}{\cellcolor{gray!15}}
\begin{tabular}{l c c c c c c c c c c c}
\toprule
& \multicolumn{6}{c}{\textbf{General Classification}} & \multicolumn{5}{c}{\textbf{Image Retrieval}} \\
\cmidrule(lr){2-7} \cmidrule(lr){8-12}
\textbf{Auxiliary loss}
& \rotatebox{90}{Avg.\ Class.}
& \rotatebox{90}{\shortstack{{\fontsize{7}{8}\selectfont ImageNet}\\{\fontsize{3}{4}\selectfont Val \cite{imagenet1k}}}}
& \rotatebox{90}{\shortstack{{\fontsize{7}{8}\selectfont ImageNet}\\{\fontsize{3}{4}\selectfont v2 \cite{imagenetv2}}}}
& \rotatebox{90}{{\fontsize{7}{8}\selectfont ObjectNet \cite{objectnet}}}
& \rotatebox{90}{\shortstack{{\fontsize{7}{8}\selectfont ImageNet}\\{\fontsize{3}{4}\selectfont Adversarial \cite{imageneta}}}}
& \rotatebox{90}{\shortstack{{\fontsize{7}{8}\selectfont ImageNet}\\{\fontsize{3}{4}\selectfont Renditions \cite{imagenetr}}}}
& \rotatebox{90}{Avg.\ Retrieval}
& \rotatebox{90}{\shortstack{{\fontsize{7}{8}\selectfont COCO}\\{\fontsize{3}{4}\selectfont T$\rightarrow$I \cite{coco}}}}
& \rotatebox{90}{\shortstack{{\fontsize{7}{8}\selectfont COCO}\\{\fontsize{3}{4}\selectfont I$\rightarrow$T \cite{coco}}}}
& \rotatebox{90}{\shortstack{{\fontsize{7}{8}\selectfont Flickr30k}\\{\fontsize{3}{4}\selectfont T$\rightarrow$I \cite{flickr}}}}
& \rotatebox{90}{\shortstack{{\fontsize{7}{8}\selectfont Flickr30k}\\{\fontsize{3}{4}\selectfont I$\rightarrow$T \cite{flickr}}}} \\
\midrule
Dense model & \gc 55.8 & 67.5 & 59.3 & 50.6 & 27.8 & 73.7 & \gc 58.4 & 36.3 & 54.4 & 63.3 & 79.6 \\
Importance and load loss \cite{vmoe} & \gc 59.9 & 70.6 & 62.3 & 54.3 & 34.7 & 77.7 & \gc 59.7 & 37.8 & 56.3 & 64.4 & 80.2 \\
Entropy loss \cite{liemoe} & \gc 59.9 & 70.7 & 62.2 & 53.9 & 35.5 & 77.1 & \gc 60.0 & 38.5 & 56.4 & 65.0 & 80.1 \\
Loss-free \cite{aux_free} & \gc 62.6 & 72.5 & 64.4 & 56.4 & 40.0 & 79.9 & \gc 60.9 & \textbf{39.2} & \textbf{56.8} & 65.3 & 82.5 \\
Ours & \gc \textbf{63.2} & \textbf{72.9} & \textbf{64.9} & \textbf{57.7} & \textbf{40.4} & \textbf{80.0} & \gc \textbf{61.4} & 38.9 & 56.6 & \textbf{66.5} & \textbf{83.5} \\
\bottomrule
\end{tabular}
\end{table}

\subsubsection{Video Finetuning} 
\begin{figure}[htbp]
  \centering
  \begin{subfigure}{0.3\textwidth}
    \centering
    \includegraphics[width=\linewidth]{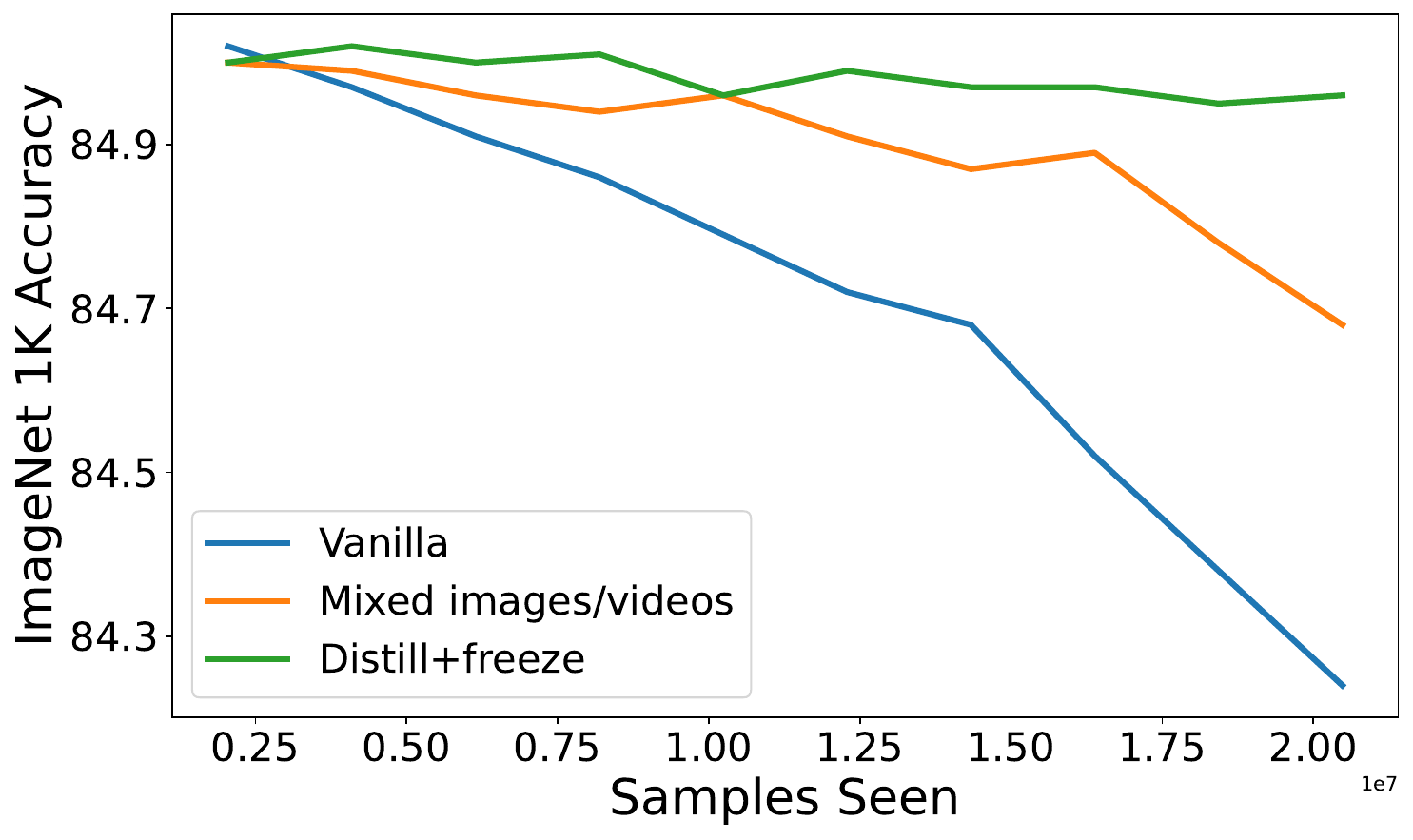}\\[0.5em]
    \includegraphics[width=\linewidth]{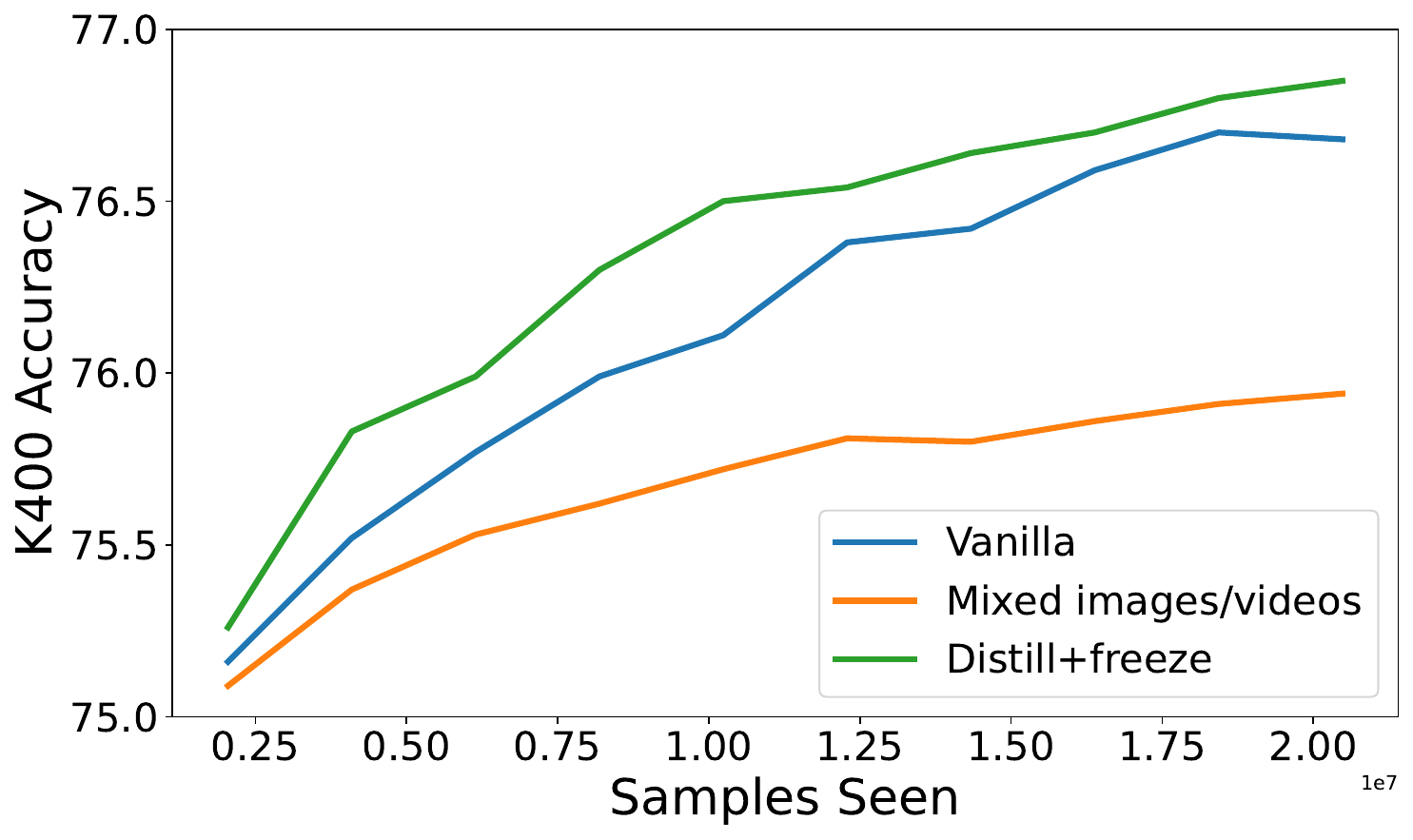}
    \caption{}
    \label{video_ft_ablation_a}
  \end{subfigure}\quad
  \begin{subfigure}{0.3\textwidth}
    \centering
    \includegraphics[width=\linewidth]{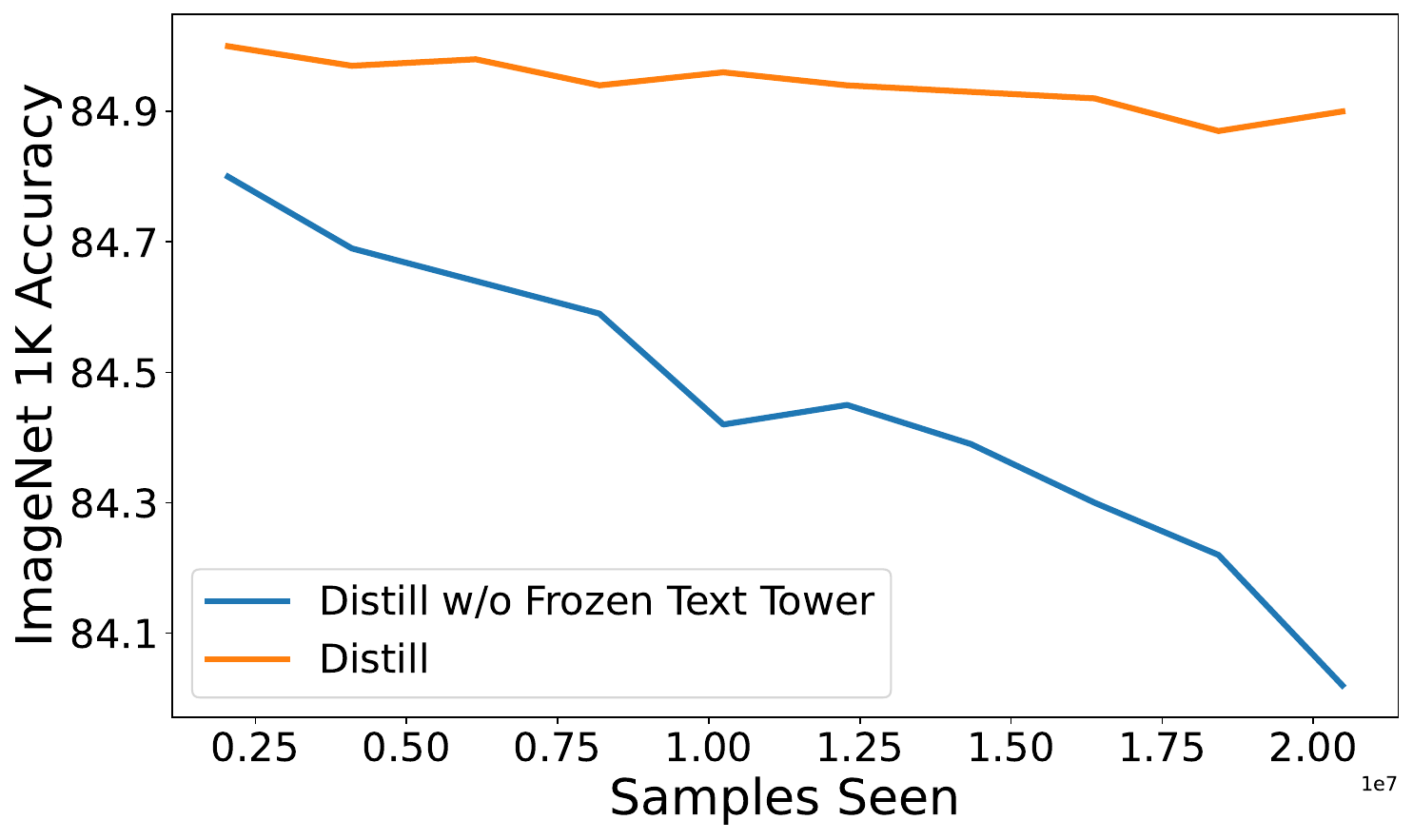}\\[0.5em]
    \includegraphics[width=\linewidth]{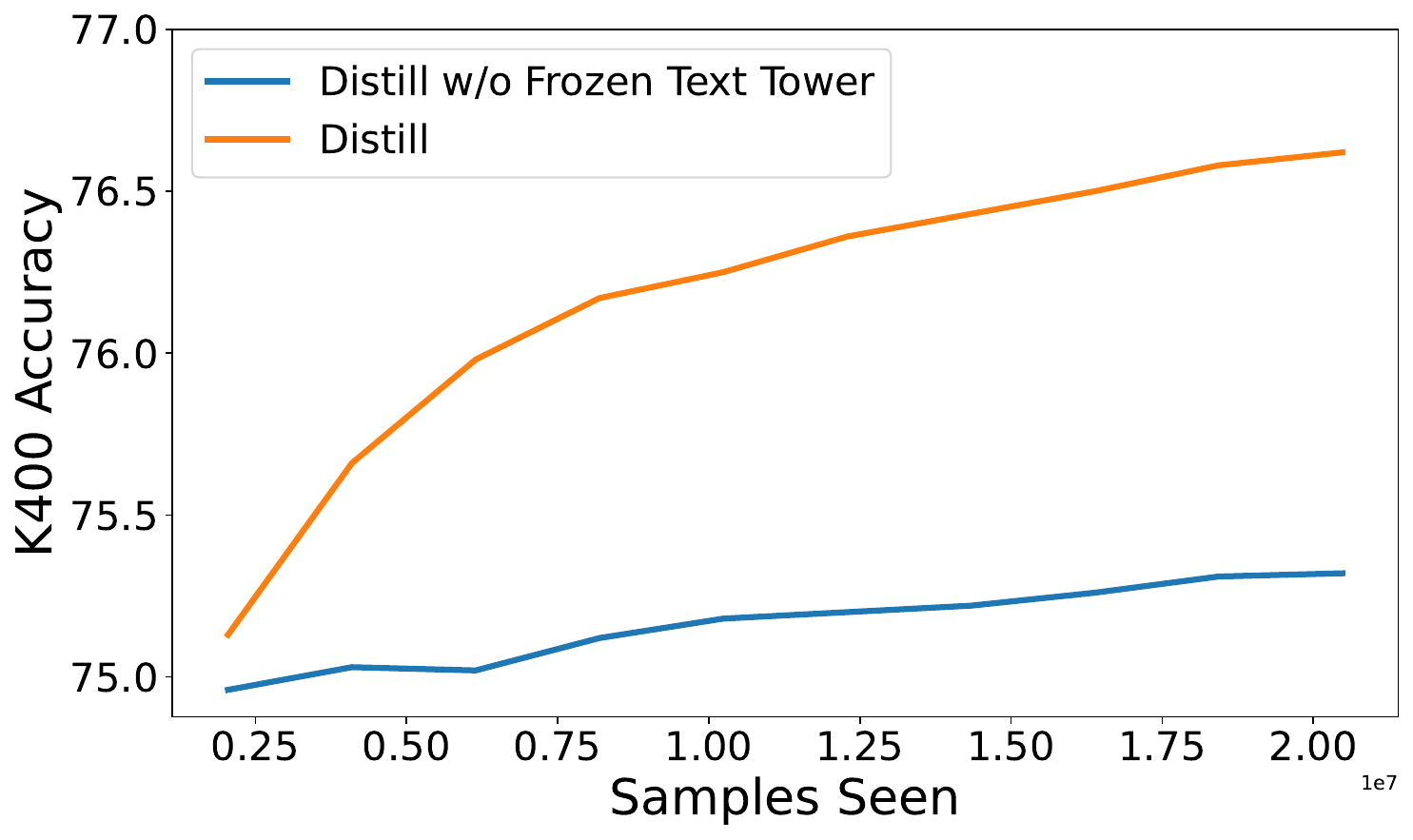}
    \caption{}
    \label{video_ft_ablation_b}
  \end{subfigure}\quad
  \begin{subfigure}{0.3\textwidth}
    \centering
    \includegraphics[width=\linewidth]{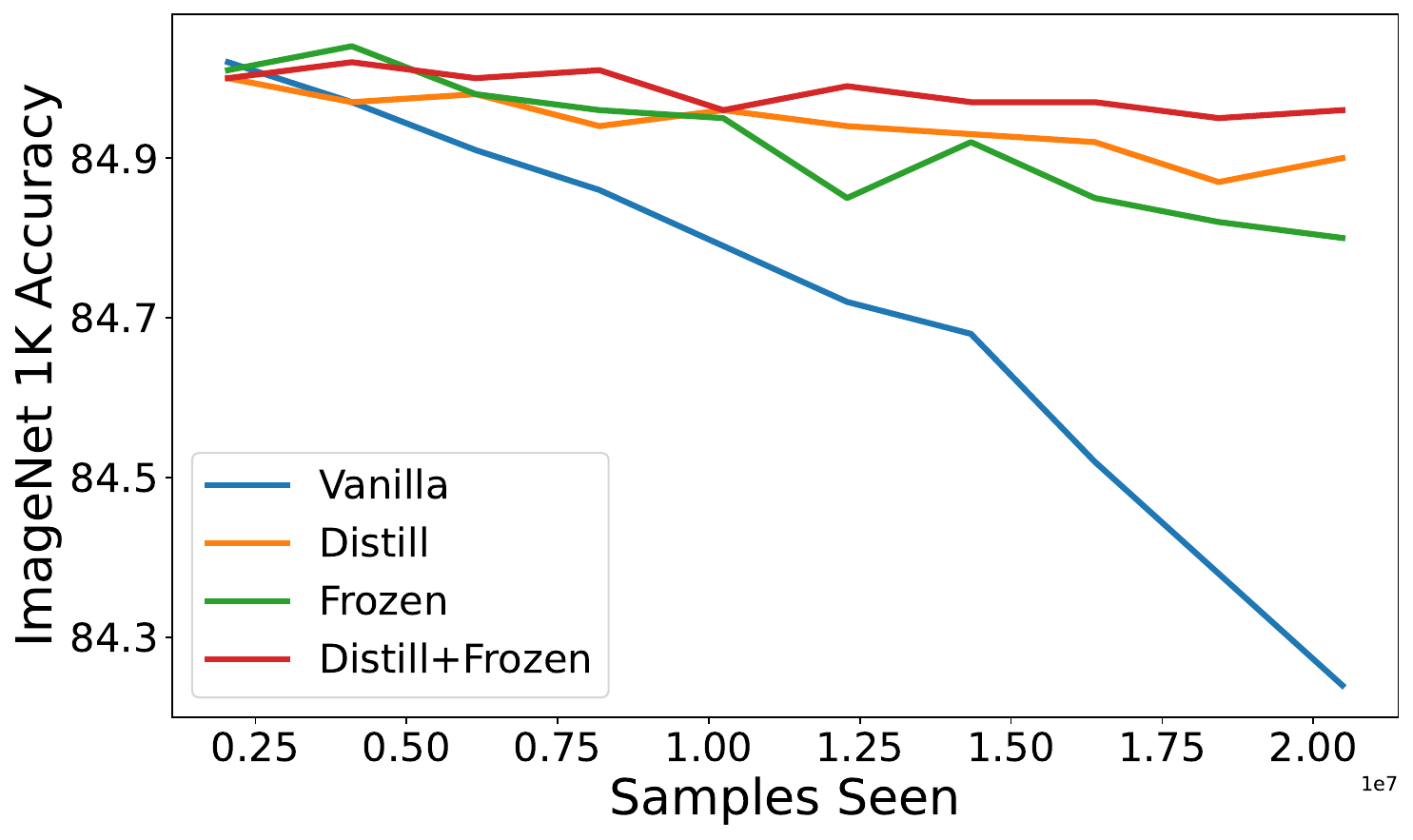}\\[0.5em]
    \includegraphics[width=\linewidth]{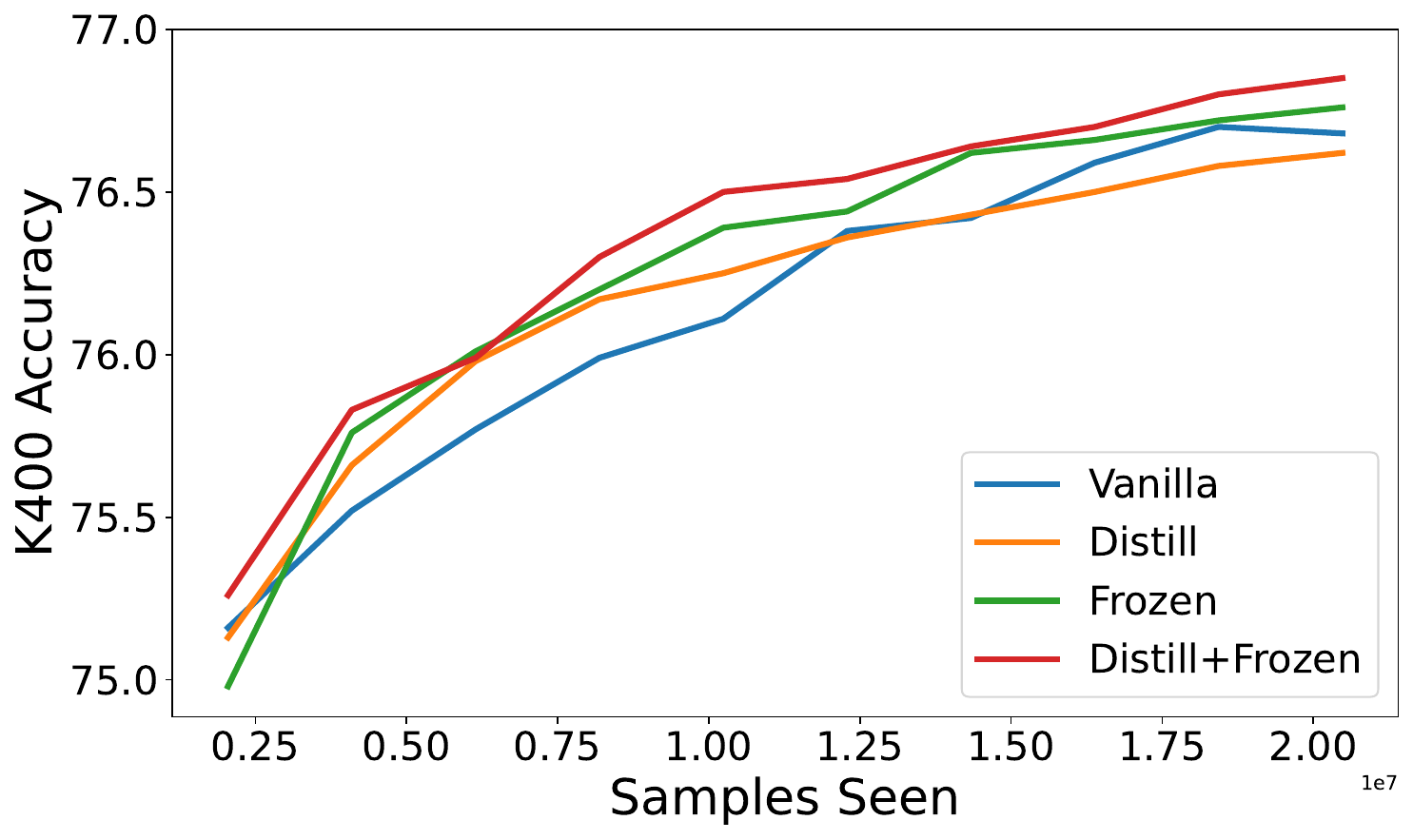}
    \caption{}
    \label{video_ft_ablation_c}
  \end{subfigure}
  \caption{Ablations of different video finetuning methods. (a) Our proposed distillation and freezing techniques outperform finetuning with vanilla video data, or mixed image and video data; (b) We observe significant degradation if we apply frame-level distillation without freezing the text tower; (c) The combination of the proposed distillation and freezing methods yields the best results.}
  \label{video_ft_ablation}
\end{figure}

We ablate the video finetuning components introduced in Section~\ref{video_ft} in Fig. \ref{video_ft_ablation}, using MoE-ViE-H. For clarity, we report ImageNet-1K as a representative image benchmark and K400 as a representative video benchmark. 

As shown in Fig. \ref{video_ft_ablation_a}, while naive video finetuning on video data improves video performance, it causes a significant degradation on image benchmarks over time. Mixing image data into the video finetuning partially mitigates this issue, but fails to prevent degradation, and limits our gains on video understanding. We find that applying frame-level distillation without freezing the MLP layers in the text tower sharply hurts image performance. Finally, we see that, individually, both distillation and expert freezing help preserve image performance, and that their combination achieves the best results on both image and video benchmarks.

\subsection{Expert Specialization}
\label{moe_analysis}
We conduct both qualitative and quantitative analysis of MoE-ViE’s expert activation patterns to better understand how fine-grained MoE captures complex, multi-faceted visual information. Across both tokens and tasks, we observe clear and meaningful expert specialization.

\begin{figure}[t]
  \centering

  \begin{subfigure}{\textwidth}
    \centering

    \begin{subfigure}{0.16\textwidth}
      \centering
      \includegraphics[height=2cm,width=\linewidth]{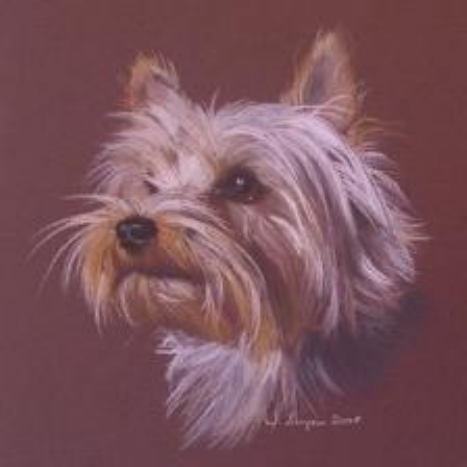}
    \end{subfigure}
    \begin{subfigure}{0.16\textwidth}
      \centering
      \includegraphics[height=2cm,width=\linewidth]{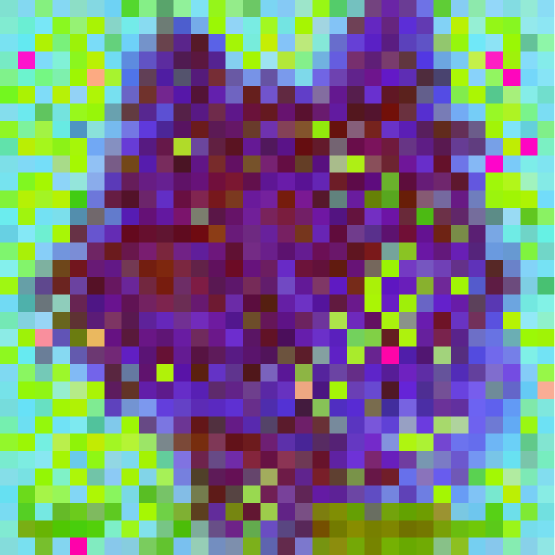}
    \end{subfigure}
    \begin{subfigure}{0.16\textwidth}
      \centering
      \includegraphics[height=2cm,width=\linewidth]{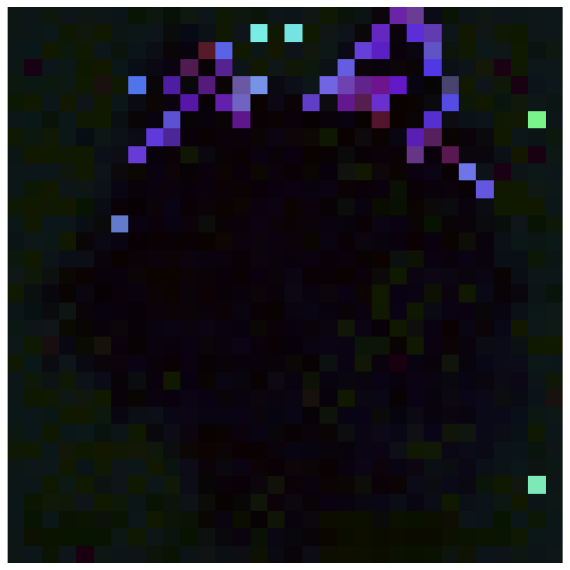}
    \end{subfigure}
    \begin{subfigure}{0.16\textwidth}
      \centering
      \includegraphics[height=2cm,width=\linewidth]{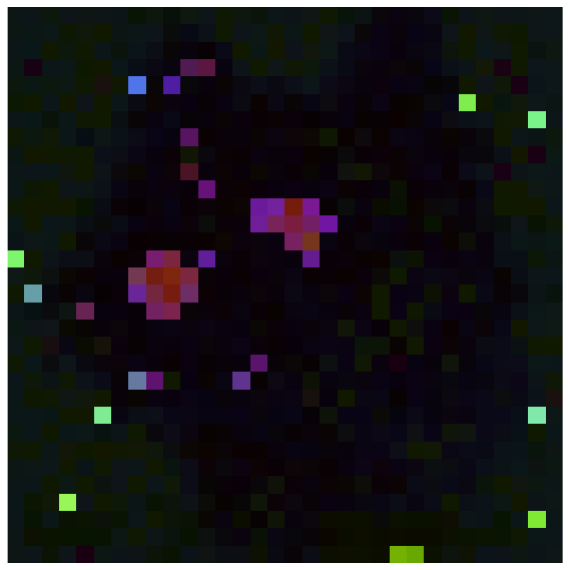}
    \end{subfigure}
    \begin{subfigure}{0.16\textwidth}
      \centering
      \includegraphics[height=2cm,width=\linewidth]{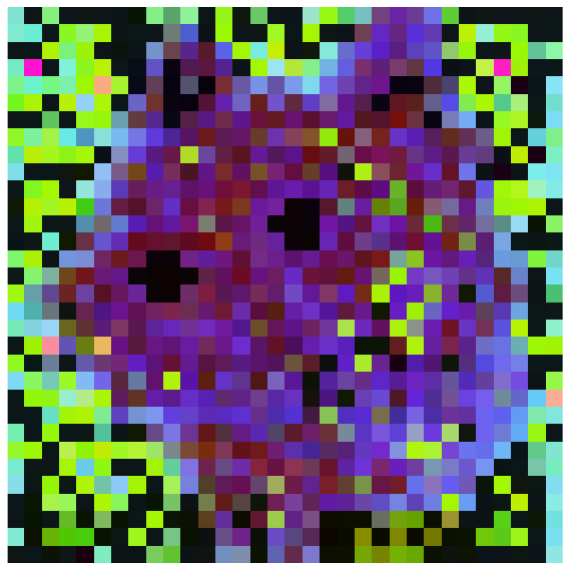}
    \end{subfigure}
    \begin{subfigure}{0.16\textwidth}
      \centering
      \includegraphics[height=2cm,width=\linewidth]{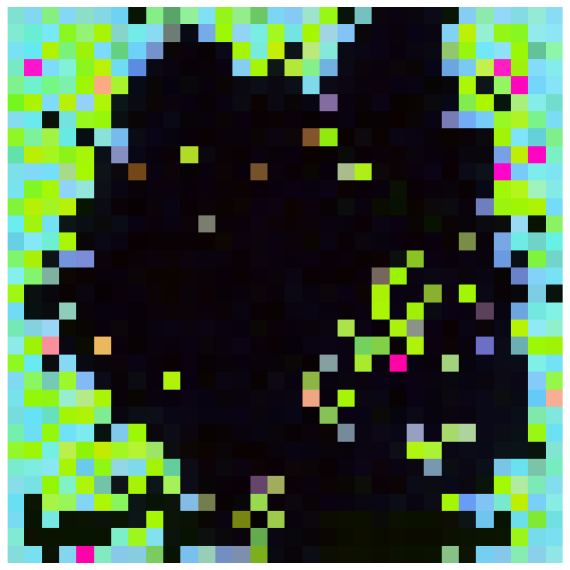}
    \end{subfigure}

    \caption{From left to right: raw image, shared expert, experts 11, 18, 22, 29}
    \label{fig:exp_examples_dog}
  \end{subfigure}

  \par\medskip

  \begin{subfigure}{\textwidth}
    \centering

    \begin{subfigure}{0.16\textwidth}
      \centering
      \includegraphics[height=2cm,width=\linewidth]{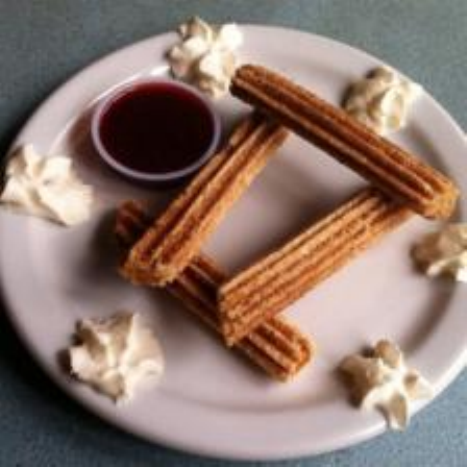}
    \end{subfigure}
    \begin{subfigure}{0.16\textwidth}
      \centering
      \includegraphics[height=2cm,width=\linewidth]{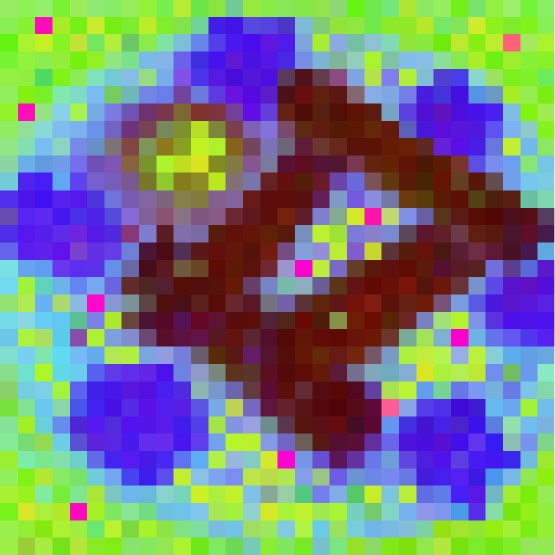}
    \end{subfigure}
    \begin{subfigure}{0.16\textwidth}
      \centering
      \includegraphics[height=2cm,width=\linewidth]{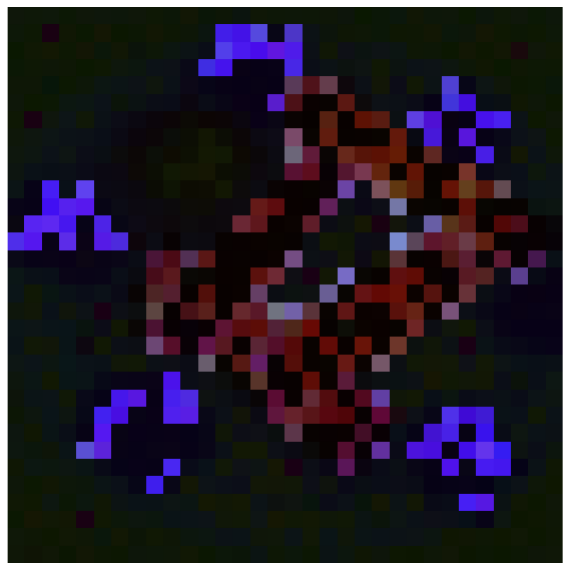}
    \end{subfigure}
    \begin{subfigure}{0.16\textwidth}
      \centering
      \includegraphics[height=2cm,width=\linewidth]{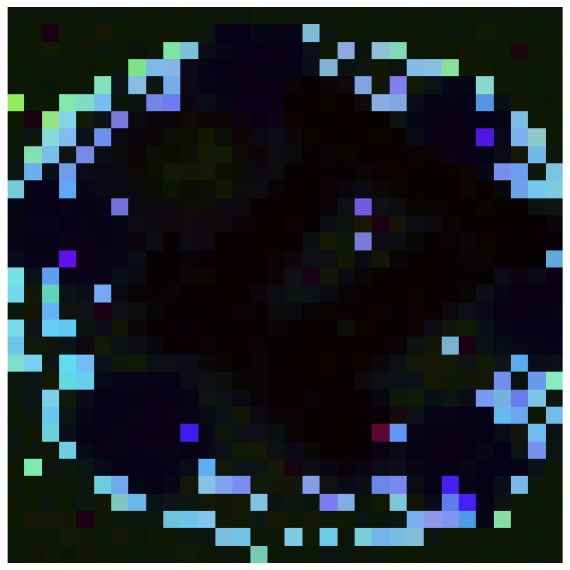}
    \end{subfigure}
    \begin{subfigure}{0.16\textwidth}
      \centering
      \includegraphics[height=2cm,width=\linewidth]{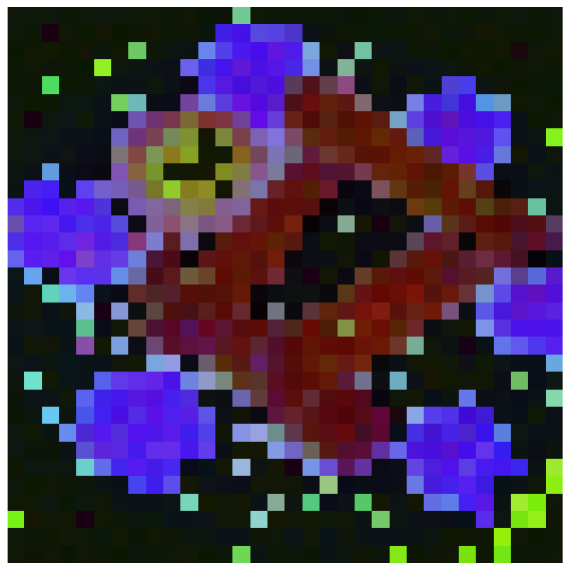}
    \end{subfigure}
    \begin{subfigure}{0.16\textwidth}
      \centering
      \includegraphics[height=2cm,width=\linewidth]{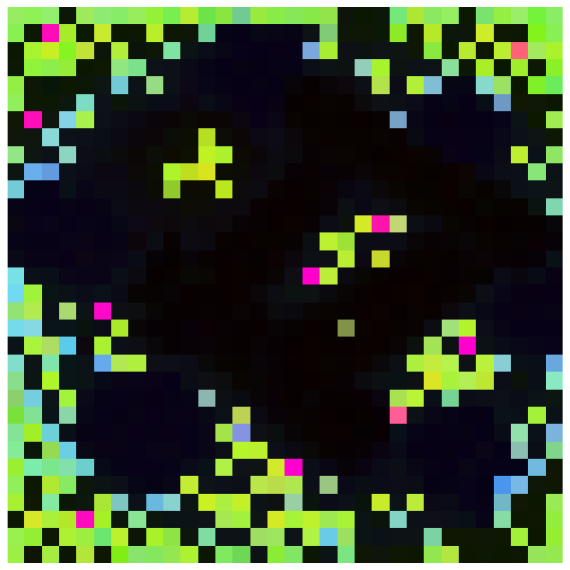}
    \end{subfigure}

    \caption{From left to right: raw image, shared expert, experts 11, 17, 19, 29}
    \label{fig:exp_examples_food}
  \end{subfigure}

  \caption{MoE Expert activations in layer 20 of MoE-ViE-H/14. Experts specialize for distinct, semantically meaningful, aspects of the image.}
  \label{fig:expert_activations_per_token}
\end{figure}

We first project each layer’s features to three principal components using PCA, and map the resulting 3D embeddings to RGB, as in \cite{pe,dinov3}. For each expert, we visualize the tokens routed to it, as shown in Fig.~\ref{fig:expert_activations_per_token}. Since we use top-$k$ routing, these subplots are not mutually exclusive, \ie, a token may appear in multiple expert-specific views. We also illustrate the subplot for the shared expert, which receives all tokens, preserving global context. 

Such visualization reveals several compelling insights: (1) each expert extracts distinct, semantically meaningful aspects of the image; (2) these specializations are complementary, such that their combined outputs yield richer visual understanding. For example, in Fig.~\ref{fig:expert_activations_per_token}\subref{fig:exp_examples_dog}, expert 11 predominantly activates on tokens corresponding to the dog’s ears, while expert 18 emphasizes the eyes and nose. In Fig.~\ref{fig:expert_activations_per_token}\subref{fig:exp_examples_food}, expert 17 highlights the plate, whereas expert 19 concentrates on the food. In both examples, expert 29 consistently activates on background regions.

\begin{figure}[htbp]
  \centering
  \newcommand{\yaxw}{0.12\textwidth}
  \newcommand{\plotw}{0.29\textwidth}
  \newcommand{\yaxscale}{1.35} 
  \newcommand{\yaxraise}{1.5ex} 
  \begin{subfigure}[b]{\yaxw}
    \centering
    \raisebox{\yaxraise}{%
      \scalebox{\yaxscale}{%
        \includegraphics[width=\linewidth]{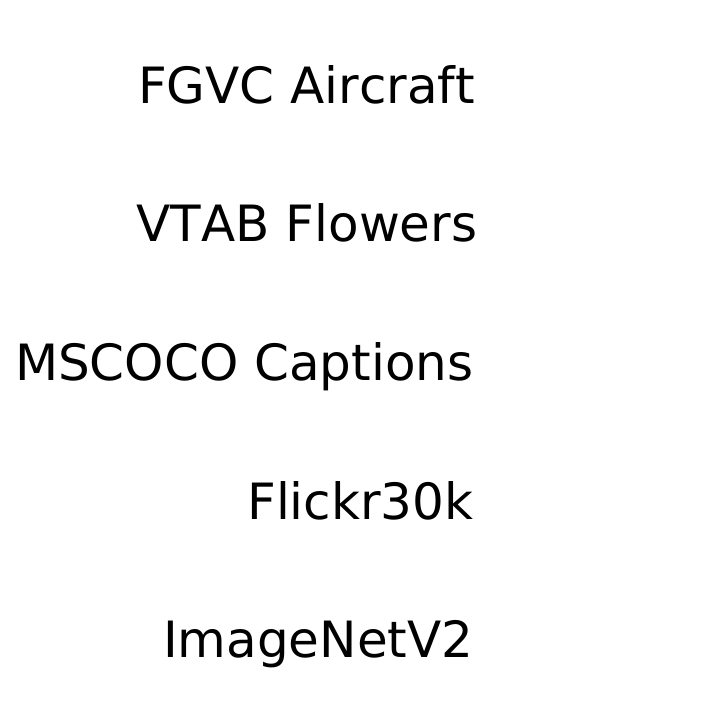}%
      }%
    }
    \caption*{}
  \end{subfigure}\hfill
  \begin{subfigure}[b]{\plotw}
    \centering
    \includegraphics[width=\linewidth]{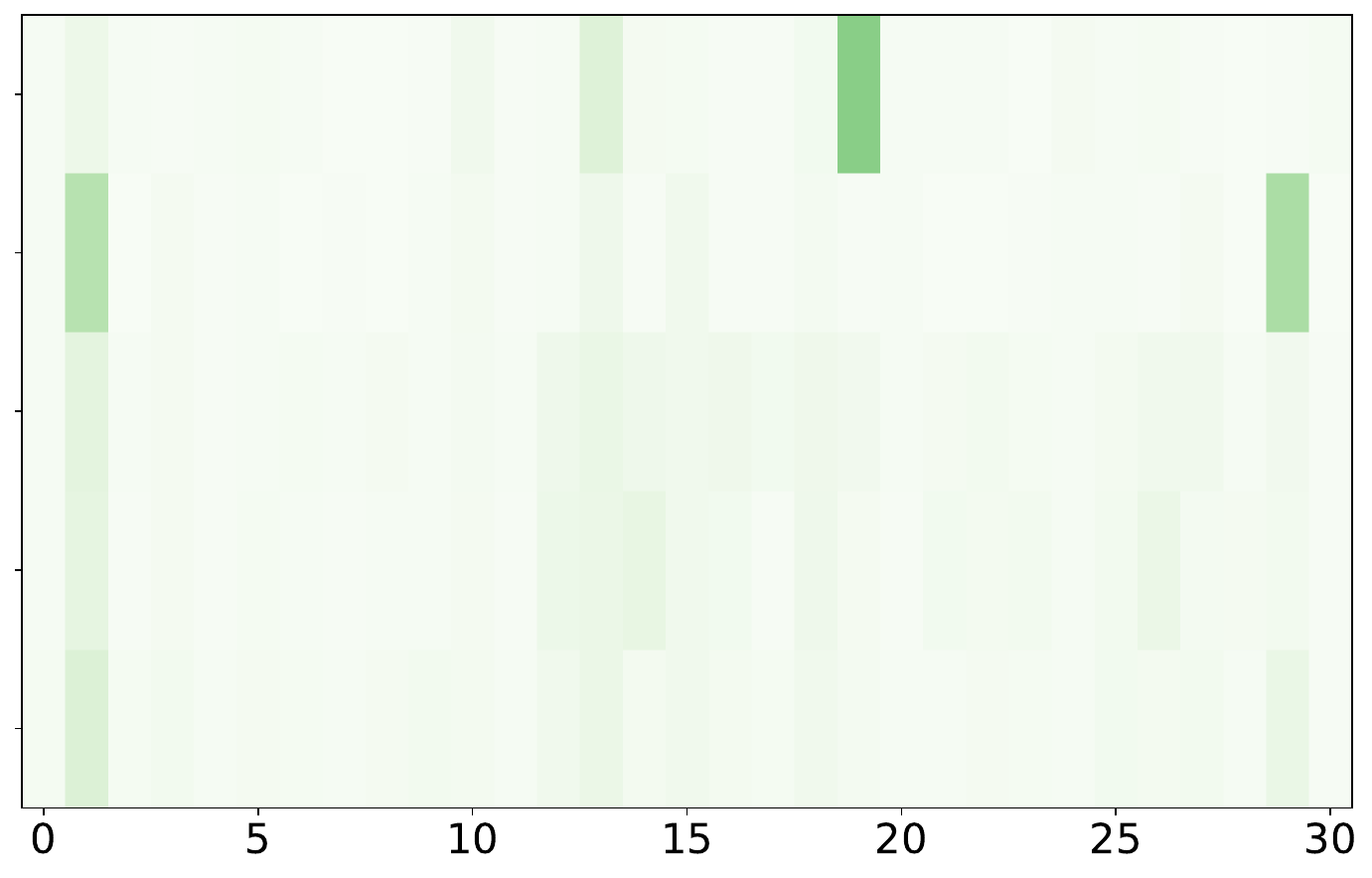}
    \caption{MoE layer 29}
  \end{subfigure}\hfill
  \begin{subfigure}[b]{\plotw}
    \centering
    \includegraphics[width=\linewidth]{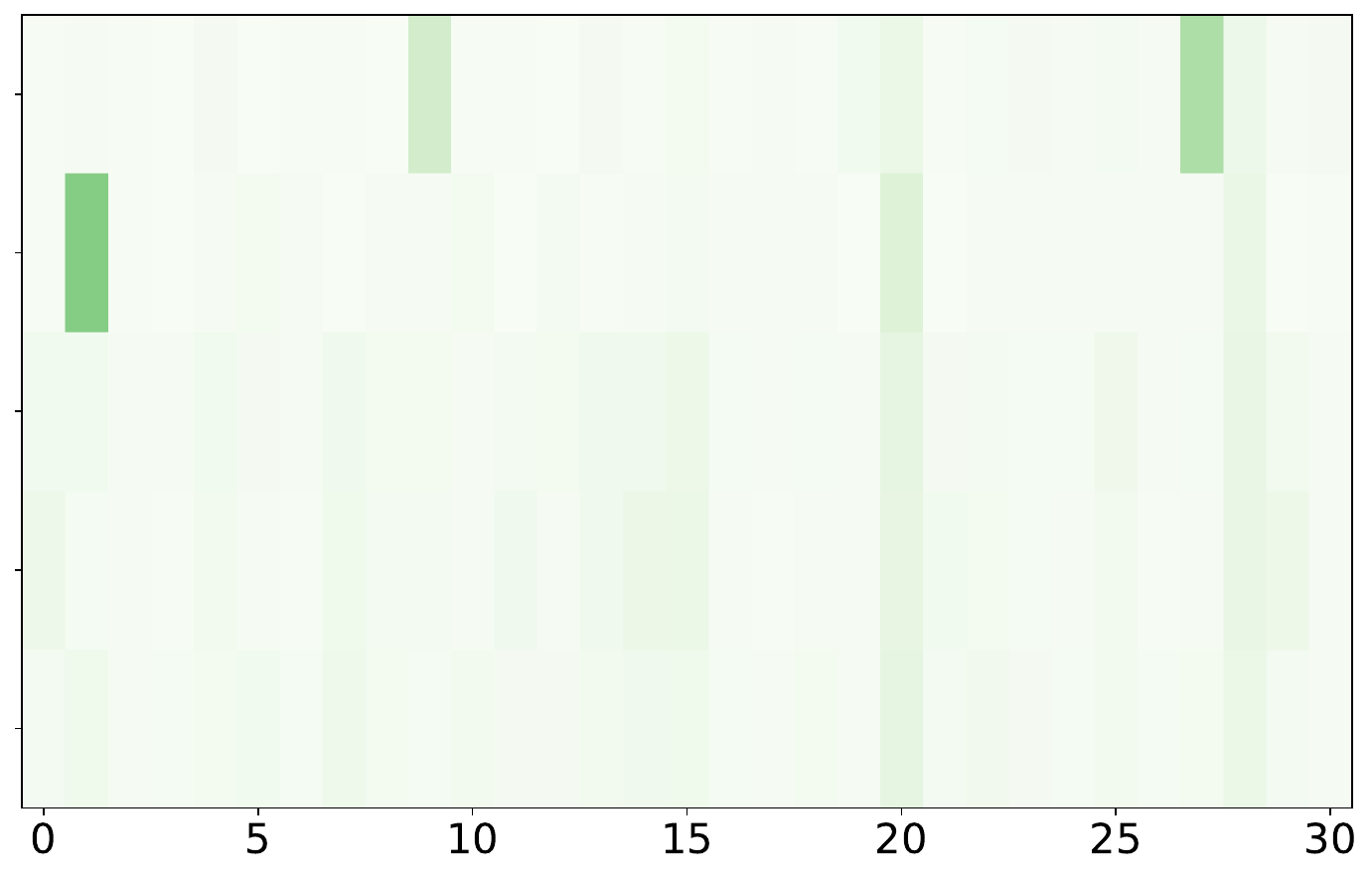}
    \caption{MoE layer 30}
  \end{subfigure}\hfill
  \begin{subfigure}[b]{\plotw}
    \centering
    \includegraphics[width=\linewidth]{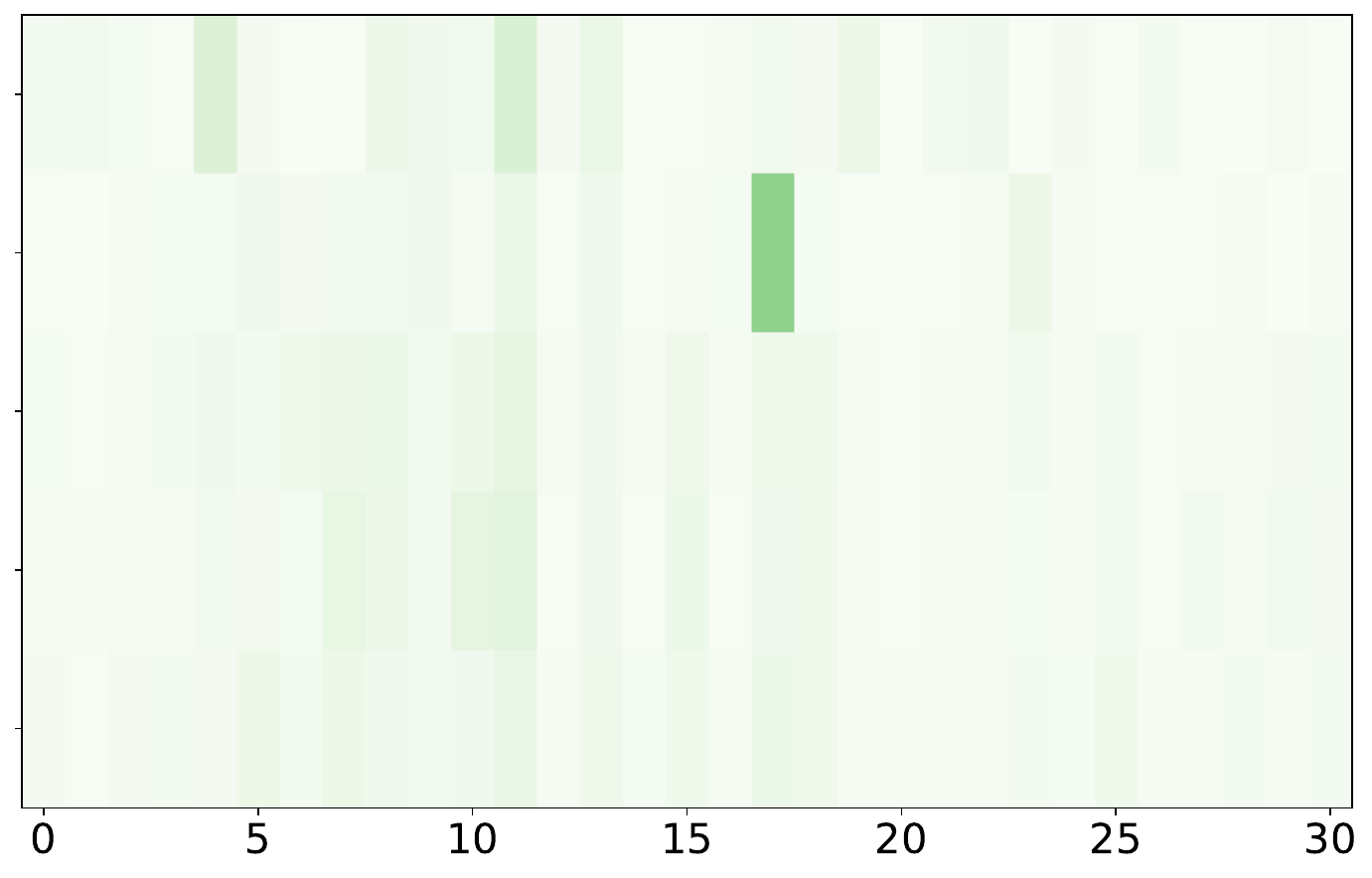}
    \caption{Last MoE layer, layer 31}
  \end{subfigure}
  \caption{Expert activation across all data for a given task, final layers of MoE-ViE-H/14.}
\label{fig:expert_activations_per_task}
\end{figure}

We further quantify expert specialization through per-task expert utilization heatmaps (Fig.~\ref{fig:expert_activations_per_task}) and routing entropy (Table~\ref{routing_entropy}). As shown in Fig.~\ref{fig:expert_activations_per_task}, experts exhibit distinct activation profiles across tasks, consistent with the emergence of task-relevant specialization in deeper MoE layers. Interestingly, on domain-specific benchmarks (\eg, FGVC Aircraft, VTAB Flowers), activations are more imbalanced across experts than on more general benchmarks such as ImageNet or COCO. This concentration may help explain MoE-ViE’s stronger performance on fine-grained image benchmarks. To complement this analysis, we measure routing entropy across tasks and layers, summarized in Table~\ref{routing_entropy}. Entropy consistently decreases with depth, confirming higher specialization at later layers. Moreover, benchmarks spanning narrower visual domains (VTAB Flowers, FGVC Aircraft) exhibit lower entropy overall, indicating stronger semantic specialization.

\begin{table}[H]
\centering
\caption{Routing entropy averaged over layer ranges.}
\label{routing_entropy}
\fontsize{6.5}{8}\selectfont
\setlength{\tabcolsep}{2.7pt}
\newcommand{\gc}{\cellcolor{gray!10}}
\begin{tabular}{lccccc}
\toprule
Benchmark & L0--5 & L6--15 & L16--25 & L26--30&Uniform \\
\midrule
VTAB Flowers \cite{flowers}   & 3.20 & 3.17 & 2.83 & 2.30 & 3.47 \\
FGVC Aircraft \cite{aircraft}  & 3.19 & 3.26 & 2.95 & 2.56 & 3.47 \\
ImageNetV2 \cite{imagenetv2}     & 3.34 & 3.31 & 3.30 & 3.21 & 3.47 \\
Flickr30k \cite{flickr}      & 3.28 & 3.30 & 3.23 & 3.07 & 3.47 \\
COCO \cite{coco}  & 3.30 & 3.32 & 3.30 & 3.18 & 3.5 \\
\bottomrule
\end{tabular}
\end{table}

\section{Related Work}

\subsection{Contrastive Language-Image Pretraining}

Contrastive vision-language pretraining (CLIP) has emerged as a foundational approach for building strong and transferable vision encoders \cite{clip,align}. A number of extensions refine either its training objective or supervision signals. For instance, SigLIP \cite{siglip} replaces the Softmax used to compute InfoNCE loss \cite{infonce} with a Sigmoid function, enabling better scaling and efficiency. MaskCLIP \cite{maskclip} augments masked image modeling to train the vision model as a self-distillation. CoCa \cite{cocca} incorporates caption loss in contrastive training, and LocCa \cite{locca} further introduces grounding supervision. SigLIP 2 \cite{siglip2} and PE \cite{pe} integrate several of these ingredients, achieving high accuracy on zero-shot image and video benchmarks. 

\subsection{Mixture-of-Experts (MoE)}

MoE scales model capacity by replacing dense layers with an expert router and a set of expert layers. The learned expert router activates a small subset of experts per input, yielding sparse input-dependent computation. The idea traces back to conditional computation with expert gating formulations \cite{early_moe_1,early_moe_2,early_moe_4}, and has since become a practical mechanism for scaling transformers via sparse top-$k$ routing \cite{original_moe}. MoE has been extensively studied in natural language processing (NLP) \cite{switch,gshard}, and has shown great success in many recent large language models \cite{qwen3,deepseekv3,gptoss,kimik2,mixtral}. Early MoE for vision efforts largely follow supervised training strategies \cite{vmoe,rmoe,adamv} or introduce MoE during the post-training stages \cite{mome,tove,omni_smola}. Later works have introduced MoE into CLIP pretraining \cite{clip}. LIMoE \cite{liemoe} pioneered contrastive learning with sparsely activated experts and demonstrated improved zero-shot image benchmarks. CLIP-MoE \cite{clip_moe} and CLIP-UP \cite{moe_clipup} introduce upcycling techniques to convert pretrained dense CLIP models to MoE variants. Despite these advances, prior MoE vision encoders have not matched performance of SOTA dense models.

\section{Conclusion}

In this work, we present MoE-ViE, an efficient and performant vision encoder with fine-grained MoE, for image and video understanding. By combining our proposed architectural design, kernel optimization, magnitude-aware loss-free load balancing control, and refined video finetuning, MoE-ViE outperforms both prior dense and MoE vision encoders across image and video benchmarks at all scales. Our largest model achieves comparable results to a SOTA dense encoder that is
$1.7\times$ larger while running at 76\% of its latency, demonstrating the significant potential for further effectively scaling up vision encoders.

\section*{Acknowledgements}
We would like to thank Giang Nguyen, Jiankun Liu, Zhuangqun Huang, Tuyen Tran, Xiao Zhang, Parisa Hassanzadeh, Lambert Mathias, Po-Yao Huang, Daniel Bolya for their contributions, support and discussions for this project.

\appendix

\section{Architecture Design Ablations}
\label{supp_a}

This section presents additional analyses of our MoE design, including our study on shared experts, activated experts, and the expert width. Unless otherwise noted, we train MoE-ViE-B with patch size 32 on 12.8B samples from MetaCLIP \cite{metaclip}.

\subsection{Number of Shared Experts}

Table~\ref{num_shared_experts} compares models with different numbers of shared experts. To keep the compute budget fixed, we hold the total number of experts constant (32 in this case) and trade off shared experts against sparsely routed experts. We see that introducing shared experts generally improves performance over a fully sparse routing configuration. However, increasing the number of shared experts beyond a small value yields no further gains and can even degrade results. Therefore, we use a single shared expert in all experiments for simplicity.

\begin{table}[htbp]
\centering
\caption{Ablation on the number of shared experts.}
\label{num_shared_experts}
\fontsize{6.5}{8}\selectfont
\setlength{\tabcolsep}{2.5pt}
\newcommand{\gc}{\cellcolor{gray!10}}
\begin{tabular}{c c c c c c c c c c c c}
\toprule
& \multicolumn{6}{c}{\textbf{General Classification}} & \multicolumn{5}{c}{\textbf{Image Retrieval}} \\
\cmidrule(lr){2-7} \cmidrule(lr){8-12}
\textbf{\# of shared experts}
& \rotatebox{90}{Avg.\ Class.}
& \rotatebox{90}{\shortstack{{\fontsize{7}{8}\selectfont ImageNet}\\{\fontsize{3}{4}\selectfont Val \cite{imagenet1k}}}}
& \rotatebox{90}{\shortstack{{\fontsize{7}{8}\selectfont ImageNet}\\{\fontsize{3}{4}\selectfont v2 \cite{imagenetv2}}}}
& \rotatebox{90}{{\fontsize{7}{8}\selectfont ObjectNet \cite{objectnet}}}
& \rotatebox{90}{\shortstack{{\fontsize{7}{8}\selectfont ImageNet}\\{\fontsize{3}{4}\selectfont Adversarial \cite{imageneta}}}}
& \rotatebox{90}{\shortstack{{\fontsize{7}{8}\selectfont ImageNet}\\{\fontsize{3}{4}\selectfont Renditions \cite{imagenetr}}}}
& \rotatebox{90}{Avg.\ Retrieval}
& \rotatebox{90}{\shortstack{{\fontsize{7}{8}\selectfont COCO}\\{\fontsize{3}{4}\selectfont T$\rightarrow$I \cite{coco}}}}
& \rotatebox{90}{\shortstack{{\fontsize{7}{8}\selectfont COCO}\\{\fontsize{3}{4}\selectfont I$\rightarrow$T \cite{coco}}}}
& \rotatebox{90}{\shortstack{{\fontsize{7}{8}\selectfont Flickr30k}\\{\fontsize{3}{4}\selectfont T$\rightarrow$I \cite{flickr}}}}
& \rotatebox{90}{\shortstack{{\fontsize{7}{8}\selectfont Flickr30k}\\{\fontsize{3}{4}\selectfont I$\rightarrow$T \cite{flickr}}}} \\
\midrule
0 &  \gc 62.1 & 72.5 & 64.2 & 56.1 & 38.4 & 79.5& \gc 61.1	& 38.8 & 56.4 & 66.2 & 83.0 \\
1 & \gc \textbf{63.2}& 72.9	&	\textbf{64.9}& 57.7&	\textbf{40.4} &	\textbf{80.0}	&\gc 	\textbf{61.4}&	38.9& \textbf{56.6}	&\textbf{66.5}	& 83.5  \\
2 &\gc63.0 & \textbf{73.2}&	64.5& \textbf{57.9} & 39.7 &	79.9 &\gc 61.3	& \textbf{39.1} & 56.4	& 66.3& \textbf{83.6} \\
4 &\gc 62.5& 72.6 & 64.2 & 57.2 & 38.8 & 79.5 &\gc 60.2 & 37.5 & 56.5 & 65.0 & 81.7 \\
\bottomrule
\end{tabular}
\end{table}

\subsection{Weight of Shared Experts}
\label{appx_shard_experts}

We sweep the scaling factor applied to the shared expert. Table \ref{weight_shared_experts} summarizes the results. Our model performance is largely insensitive to this weight across a broad range, suggesting that it is not a critical hyperparameter. We hypothesize that the scaling mainly affects the initial state. Further training can always adjust model parameters to reach a comparable solution.

\begin{table}[htbp]
\centering
\caption{Ablation on the weight of shared experts.}
\label{weight_shared_experts}
\fontsize{6.5}{8}\selectfont
\setlength{\tabcolsep}{2.5pt}
\newcommand{\gc}{\cellcolor{gray!10}}
\begin{tabular}{l c c c c c c c c c c c}
\toprule
& \multicolumn{6}{c}{\textbf{General Classification}} & \multicolumn{5}{c}{\textbf{Image Retrieval}} \\
\cmidrule(lr){2-7} \cmidrule(lr){8-12}
\textbf{$\lambda$}
& \rotatebox{90}{Avg.\ Class.}
& \rotatebox{90}{\shortstack{{\fontsize{7}{8}\selectfont ImageNet}\\{\fontsize{3}{4}\selectfont Val \cite{imagenet1k}}}}
& \rotatebox{90}{\shortstack{{\fontsize{7}{8}\selectfont ImageNet}\\{\fontsize{3}{4}\selectfont v2 \cite{imagenetv2}}}}
& \rotatebox{90}{{\fontsize{7}{8}\selectfont ObjectNet \cite{objectnet}}}
& \rotatebox{90}{\shortstack{{\fontsize{7}{8}\selectfont ImageNet}\\{\fontsize{3}{4}\selectfont Adversarial \cite{imageneta}}}}
& \rotatebox{90}{\shortstack{{\fontsize{7}{8}\selectfont ImageNet}\\{\fontsize{3}{4}\selectfont Renditions \cite{imagenetr}}}}
& \rotatebox{90}{Avg.\ Retrieval}
& \rotatebox{90}{\shortstack{{\fontsize{7}{8}\selectfont COCO}\\{\fontsize{3}{4}\selectfont T$\rightarrow$I \cite{coco}}}}
& \rotatebox{90}{\shortstack{{\fontsize{7}{8}\selectfont COCO}\\{\fontsize{3}{4}\selectfont I$\rightarrow$T \cite{coco}}}}
& \rotatebox{90}{\shortstack{{\fontsize{7}{8}\selectfont Flickr30k}\\{\fontsize{3}{4}\selectfont T$\rightarrow$I \cite{flickr}}}}
& \rotatebox{90}{\shortstack{{\fontsize{7}{8}\selectfont Flickr30k}\\{\fontsize{3}{4}\selectfont I$\rightarrow$T \cite{flickr}}}} \\
\midrule
10 &  \gc 63.0 & 72.5 & 64.7 & 57.4 & \textbf{40.5} & \textbf{80.1} & \gc 61.3 & 38.6 & 56.2 & \textbf{66.8} & \textbf{83.5} \\
1 & \gc \textbf{63.2}& \textbf{72.9}	&	\textbf{64.9}& 57.7 &	40.4 &	\textbf{80.0}	&\gc 	\textbf{61.4}&	38.9& \textbf{56.6}	&66.5	& \textbf{83.5}  \\
0.1 &\gc 63.1 & 73.0	& 64.5 & \textbf{57.8} &	40.1 & 79.8	&\gc 61.1 & \textbf{39.1}	& 56.3	& 66.3 & 82.8 \\
0.01 &\gc 62.9& 72.6 & 64.5 & 57.2 & 40.4 & 80.0 &\gc 61.1	& 38.6 & 56.4 & 66.7 & 83.0 \\
\bottomrule
\end{tabular}
\end{table}

\subsection{Choice of Top-$k$}
We study the effect of the number of activated experts $k$, summarized in Table~\ref{top_k}, using a total of 32 fine-grained experts. Performance improves only marginally beyond $k=8$, indicating diminishing returns from additional expert activation. Based on this observation, we set $k\in \{4,8\}$ depending on model size and compute budget, striking a favorable balance between capacity and efficiency.

\begin{table}[htbp]
\centering
\caption{Ablations of top-$k$.}
\label{top_k}
\fontsize{6.5}{8}\selectfont
\setlength{\tabcolsep}{2.5pt}
\newcommand{\gc}{\cellcolor{gray!10}}
\begin{tabular}{l c c c c c c c c c c c}
\toprule
& \multicolumn{6}{c}{\textbf{General Classification}} & \multicolumn{5}{c}{\textbf{Image Retrieval}} \\
\cmidrule(lr){2-7} \cmidrule(lr){8-12}
\textbf{$k$}
& \rotatebox{90}{Avg.\ Class.}
& \rotatebox{90}{\shortstack{{\fontsize{7}{8}\selectfont ImageNet}\\{\fontsize{3}{4}\selectfont Val \cite{imagenet1k}}}}
& \rotatebox{90}{\shortstack{{\fontsize{7}{8}\selectfont ImageNet}\\{\fontsize{3}{4}\selectfont v2 \cite{imagenetv2}}}}
& \rotatebox{90}{{\fontsize{7}{8}\selectfont ObjectNet \cite{objectnet}}}
& \rotatebox{90}{\shortstack{{\fontsize{7}{8}\selectfont ImageNet}\\{\fontsize{3}{4}\selectfont Adversarial \cite{imageneta}}}}
& \rotatebox{90}{\shortstack{{\fontsize{7}{8}\selectfont ImageNet}\\{\fontsize{3}{4}\selectfont Renditions \cite{imagenetr}}}}
& \rotatebox{90}{Avg.\ Retrieval}
& \rotatebox{90}{\shortstack{{\fontsize{7}{8}\selectfont COCO}\\{\fontsize{3}{4}\selectfont T$\rightarrow$I \cite{coco}}}}
& \rotatebox{90}{\shortstack{{\fontsize{7}{8}\selectfont COCO}\\{\fontsize{3}{4}\selectfont I$\rightarrow$T \cite{coco}}}}
& \rotatebox{90}{\shortstack{{\fontsize{7}{8}\selectfont Flickr30k}\\{\fontsize{3}{4}\selectfont T$\rightarrow$I \cite{flickr}}}}
& \rotatebox{90}{\shortstack{{\fontsize{7}{8}\selectfont Flickr30k}\\{\fontsize{3}{4}\selectfont I$\rightarrow$T \cite{flickr}}}} \\
\midrule
4  &\gc 62.4 & 72.6 & 64.3 & 56.9 & 38.0 & 80.0 &\gc 61.0 & 38.4 & 56.0 & 66.3 & 83.2 \\
8  &\gc 63.2 & 72.9 & 64.9 & 57.7 & 40.4 & 80.0 & \gc 61.4 & 38.9 & 56.6 & 66.5 & 83.5 \\
16 &\gc 63.1 & 73.0 & 64.8 & 57.5 & 40.3 & 79.8 &\gc 61.4 & 39.1 & 56.5 & 66.3 & 83.6 \\
\bottomrule
\end{tabular}
\end{table}

\subsection{Expert Width}
\label{appx_moe_scaling}
We sweep the expert width factor $c$ (\ie, each expert has $\frac{1}{c}\times$ the dense MLP width) to study the effect of expert granularity while keeping the total expert capacity fixed at $8\times$ the MLP size. Finer granularity consistently improves performance across all benchmarks, consistent with our qualitative and quantitative analysis in Section~\ref{moe_analysis}. However, gains saturate below $\frac{1}{4}$ width, suggesting a practical lower bound on useful expert decomposition.

\begin{table}[htbp]
\centering
\caption{Ablations of expert width factor $c$.}
\label{expert_width}
\fontsize{6.5}{8}\selectfont
\setlength{\tabcolsep}{2.5pt}
\newcommand{\gc}{\cellcolor{gray!10}}
\begin{tabular}{l c c c c c c c c c c c}
\toprule
& \multicolumn{6}{c}{\textbf{General Classification}} & \multicolumn{5}{c}{\textbf{Image Retrieval}} \\
\cmidrule(lr){2-7} \cmidrule(lr){8-12}
\textbf{$c$}
& \rotatebox{90}{Avg.\ Class.}
& \rotatebox{90}{\shortstack{{\fontsize{7}{8}\selectfont ImageNet}\\{\fontsize{3}{4}\selectfont Val \cite{imagenet1k}}}}
& \rotatebox{90}{\shortstack{{\fontsize{7}{8}\selectfont ImageNet}\\{\fontsize{3}{4}\selectfont v2 \cite{imagenetv2}}}}
& \rotatebox{90}{{\fontsize{7}{8}\selectfont ObjectNet \cite{objectnet}}}
& \rotatebox{90}{\shortstack{{\fontsize{7}{8}\selectfont ImageNet}\\{\fontsize{3}{4}\selectfont Adversarial \cite{imageneta}}}}
& \rotatebox{90}{\shortstack{{\fontsize{7}{8}\selectfont ImageNet}\\{\fontsize{3}{4}\selectfont Renditions \cite{imagenetr}}}}
& \rotatebox{90}{Avg.\ Retrieval}
& \rotatebox{90}{\shortstack{{\fontsize{7}{8}\selectfont COCO}\\{\fontsize{3}{4}\selectfont T$\rightarrow$I \cite{coco}}}}
& \rotatebox{90}{\shortstack{{\fontsize{7}{8}\selectfont COCO}\\{\fontsize{3}{4}\selectfont I$\rightarrow$T \cite{coco}}}}
& \rotatebox{90}{\shortstack{{\fontsize{7}{8}\selectfont Flickr30k}\\{\fontsize{3}{4}\selectfont T$\rightarrow$I \cite{flickr}}}}
& \rotatebox{90}{\shortstack{{\fontsize{7}{8}\selectfont Flickr30k}\\{\fontsize{3}{4}\selectfont I$\rightarrow$T \cite{flickr}}}} \\
\midrule
1 & \gc 59.9& 70.6 & 62.3 & 54.3 & 34.7 & 77.7& \gc 59.7& 37.8 & 56.3 & 64.4 & 80.2 \\
2 & 61.0 & 71.1 & 62.6 & 55.0 & 37.2 & 78.9 & 60.3 & 38.0 & 56.4 & 64.9 & 82.0 \\
4 & \gc 63.2 & 72.9 & 64.9 & 57.7 & 40.4 & 80.0 & \gc 61.4 & 38.9 & 56.6 & 66.5 & 83.5 \\
8 & 62.8 & 72.7 & 64.2 & 57.9 & 39.6 & 79.5 & 61.0 & 39.0 & 56.1 & 65.9 & 83.1 \\
\bottomrule
\end{tabular}
\end{table}

\section{MoE Kernel}
\label{appx_kernel}

\subsection{Limitation of Vanilla MoE Implementation}

The vanilla implementation of MoE processes one expert at a time in a Python \texttt{for} loop. While straightforward to implement, it has
several performance limitations. First, processing experts sequentially fragments what could be a single large GEMM into many smaller, often highly imbalanced GEMMs. These small GEMMs exhibit low arithmetic intensity and insufficient parallelism, leading to poor GPU utilization and higher latency. Further, dynamic routing requires per-expert index selection to identify routed tokens, followed by separate gather/GEMM/activation/scatter launches for each expert. The resulting CPU-GPU synchronization and kernel-launch overhead can dominate latency when each expert processes only a small number of tokens. In addition, each iteration of the loop materializes multiple intermediate tensors, causing repeated reads and writes to HBM at each step. Moreover, the input is re-gathered per active expert (up to $K$ times under top-$K$ routing), creating redundant memory traffic that often makes bandwidth, instead of compute, the latency bottleneck.

\subsection{Optimized MoE Kernels}
\label{app:fused}

As described in Section~\ref{optimization}, we implement specialized kernels to address each of the above bottlenecks. The optimized implementation (Algorithm~\ref{alg:fused}) uses two custom GPU kernels written in
Triton~\cite{triton}, a compiler for writing efficient GPU programs
in Python. Algorithm~\ref{alg:fused} details the resulting procedure. The two optimizations mentioned in Section~\ref{optimization} map to the algorithm as
follows:

\begin{enumerate}
  \item \textbf{Grouped GEMM}. 
    Instead of issuing multiple separate small GEMMs
    in a loop, the optimized kernel fuses all per-expert GEMMs into a single grouped operation executed in one GPU launch, using a 3D grid that parallelizes over tokens, output columns, and experts to improve arithmetic intensity and GPU utilization. A jagged layout handles variable per-expert token counts without padding, and GPU-side \textsc{Sort}/\textsc{CumSum} replaces per-expert \textsc{Where} queries, removing the loop and CPU-GPU synchronization so the forward pass completes in only two kernel launches.

  \item \textbf{Kernel Fusion}. 
    Each kernel fuses multiple steps to reduce HBM traffic. Kernel~1 computes the gate and up projections together with the SwiGLU activation while keeping intermediates in registers, materializing only the activation output. This roughly
    halves memory traffic compared to the vanilla implementation. It also folds token gathering into the GEMM via pre-sorted indices, and Kernel~2 similarly fuses the down projection with routing-weight scaling and the output scatter (via atomic adds), writing directly to $\mathbf{Y}$ without intermediate buffers.

\end{enumerate}

\begin{algorithm}[h]
\caption{Fused MoE FFN -- Optimized Triton Kernel}
\begin{scriptsize} 
\label{alg:fused}
\begin{algorithmic}[1]
\Require Input tokens $\mathbf{X}\in\mathbb{R}^{B\times D}$; expert weights
         $\{\mathbf{W}_1^{(e)},\mathbf{W}_3^{(e)},\mathbf{W}_2^{(e)}\}_{e=1}^{E}$;
         routing indices $\mathbf{T}_{\text{id}}\in\mathbb{Z}^{B\times K}$;
         routing weights $\mathbf{T}_w\in\mathbb{R}^{B\times K}$
\Ensure  Output $\mathbf{Y}\in\mathbb{R}^{B\times D}$

\Statex \textbf{--- Phase 1: Route tokens to experts (In-parallel routing) ---}
\State $(\text{expert}[n],\;\text{pos}[n]) \gets \textsc{Sort}(\mathbf{T}_{\text{id}}.\text{flatten}())$
    \Comment{Sort by expert ID}
\State $\text{index}[n] \gets \text{pos}[n] \,/\!/ \;K$
    \Comment{Map sorted position back to original token index}
\State $\text{offsets}[e] \gets \textsc{CumSum}(\textsc{BinCount}(\text{expert},\, E))$
    \Comment{$[E\!+\!1]$; jagged segment boundaries}
\State $\text{max\_seg} \gets \max_{e}(\text{offsets}[e\!+\!1] - \text{offsets}[e])$
    \Comment{Longest expert segment}

\Statex
\Statex \textbf{--- Phase 2: Fused Gate$+$Up projection with SwiGLU (Kernel 1) ---}
\Statex \Comment{All $E$ experts execute \textbf{in parallel} as a single batched kernel launch}
\For{each GPU thread block $(m,\, n,\, e)$ \textbf{in parallel}}
    \Comment{3D grid: $\lceil\text{max\_seg}/B_M\rceil \times \lceil D'/B_N\rceil \times E$}
    \State $s \gets \text{offsets}[e]$;\enspace $l \gets \text{offsets}[e\!+\!1] - s$
    \If{$m \cdot B_M \geq l$} \textbf{return} \Comment{Skip blocks beyond this expert's segment}
    \EndIf
    \State Load tile of input rows from $\mathbf{X}$ via $\text{index}[s .. s\!+\!l]$
         \Comment{Indirect gather, fused into GEMM}
    \State $\mathbf{G}_{\text{tile}} \gets \mathbf{0}$;\enspace $\mathbf{U}_{\text{tile}} \gets \mathbf{0}$
    \For{$k = 0,\, B_K,\, 2B_K,\, \dots$}
         \Comment{Tiled matrix multiply over dimension $D$}
        \State $\mathbf{G}_{\text{tile}} \mathrel{+}= \text{tile}(\mathbf{X}) \cdot \text{tile}(\mathbf{W}_1^{(e)})$
        \State $\mathbf{U}_{\text{tile}} \mathrel{+}= \text{tile}(\mathbf{X}) \cdot \text{tile}(\mathbf{W}_3^{(e)})$
    \EndFor
    \State $\mathbf{A}_{\text{tile}} \gets \text{SiLU}(\mathbf{G}_{\text{tile}}) \odot \mathbf{U}_{\text{tile}}$
         \Comment{SwiGLU fused in registers, no memory write for $\mathbf{G}, \mathbf{U}$}
    \State Write $\mathbf{A}_{\text{tile}}$ to activation buffer
\EndFor

\Statex
\Statex \textbf{--- Phase 3: Fused Down projection$+$Scatter-Add (Kernel 2) ---}
\State $\mathbf{Y} \gets \mathbf{0}_{B\times D}$
\For{each GPU thread block $(m,\, n,\, e)$ \textbf{in parallel}}
    \Comment{3D grid over segments and output tiles}
    \State $s \gets \text{offsets}[e]$;\enspace $l \gets \text{offsets}[e\!+\!1] - s$
    \If{block's $m$-offset $\geq l$} \textbf{return} \EndIf
    \State $\mathbf{O}_{\text{tile}} \gets \mathbf{0}$
    \For{$k = 0,\, B_K,\, 2B_K,\, \dots$}
        \State $\mathbf{O}_{\text{tile}} \mathrel{+}= \text{tile}(\mathbf{A}) \cdot \text{tile}(\mathbf{W}_2^{(e)})$
    \EndFor
    \State $\mathbf{O}_{\text{tile}} \gets \mathbf{O}_{\text{tile}} \cdot \mathbf{T}_w[\text{corresponding positions}]$
         \Comment{Multiply by routing weight}
    \State $\textsc{AtomicAdd}\bigl(\mathbf{Y}[\text{index}[\cdot]],\;\mathbf{O}_{\text{tile}}\bigr)$
         \Comment{Scatter-add directly to output}
\EndFor
\State \Return $\mathbf{Y}$
\end{algorithmic}
\end{scriptsize} 
\end{algorithm}

We further utilize Grouped GEMM for training. Unlike the inference
kernel, the training kernel does not fuse activation functions into the
GEMM.  Instead, the SwiGLU activation is applied as a separate
PyTorch operator between the up projection and down projection GEMMs.  This
decoupled design allows us to flexibly experiment with different activation
types (\eg, SwiGLU, GeLU, SiLU) without modifying or re-autotuning the
Triton kernel.

\subsection{Summary}
Grouped GEMM and Kernel Fusion are our necessary implementation
mechanisms to faithfully translate MoE's algorithmic efficiency into
practical, hardware-efficient execution. This is an important implementation piece for vision encoders, which process all image patches simultaneously. With high resolution and tiling, a single image can produce tens of thousands of patch tokens, and batched inference further multiplies this by the batch size. Our kernel is designed for this
regime. The Grouped GEMM formulation maximizes compute utilization across the large token volume, while the jagged-segment layout with per-expert early-exit naturally accommodates the uneven routing distributions that arise from spatially correlated image patches.

By co-designing the MoE architecture with kernel-level optimizations that
realize its efficiency gains, we reduce vision encoder latency while
maintaining strong encoding quality, making the approach particularly
effective for end-to-end VLM applications.

\section{Scaling Experts in MoE}
\label{moe_scaling}

We conduct additional experiments to study the effect of scaling the number of experts. We train MoE-ViE-B with patch size 32 on 10.9B samples from MetaCLIP \cite{metaclip}. Across all settings, we keep the number of activated experts fixed (\ie, $k=8$) and set each expert MLP to $1/4$ of the corresponding dense MLP width.

\begin{figure}[ht]
\centering
\begin{subfigure}[b]{0.45\textwidth}
    \centering
    \includegraphics[width=\textwidth]{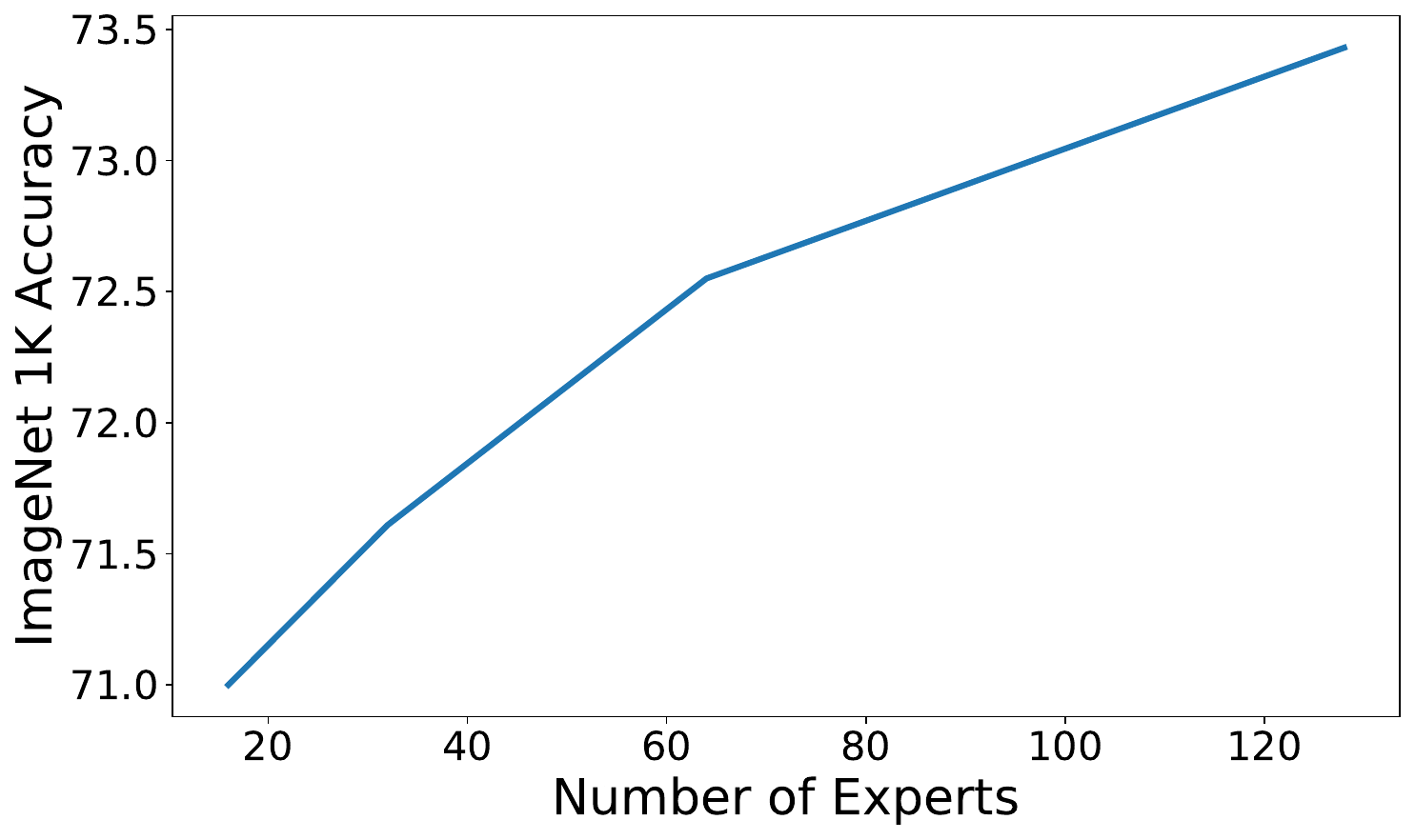}
    \caption{ImageNet 1K}
    \label{fig:sub1}
\end{subfigure}
\hfill
\begin{subfigure}[b]{0.45\textwidth}
    \centering
    \includegraphics[width=\textwidth]{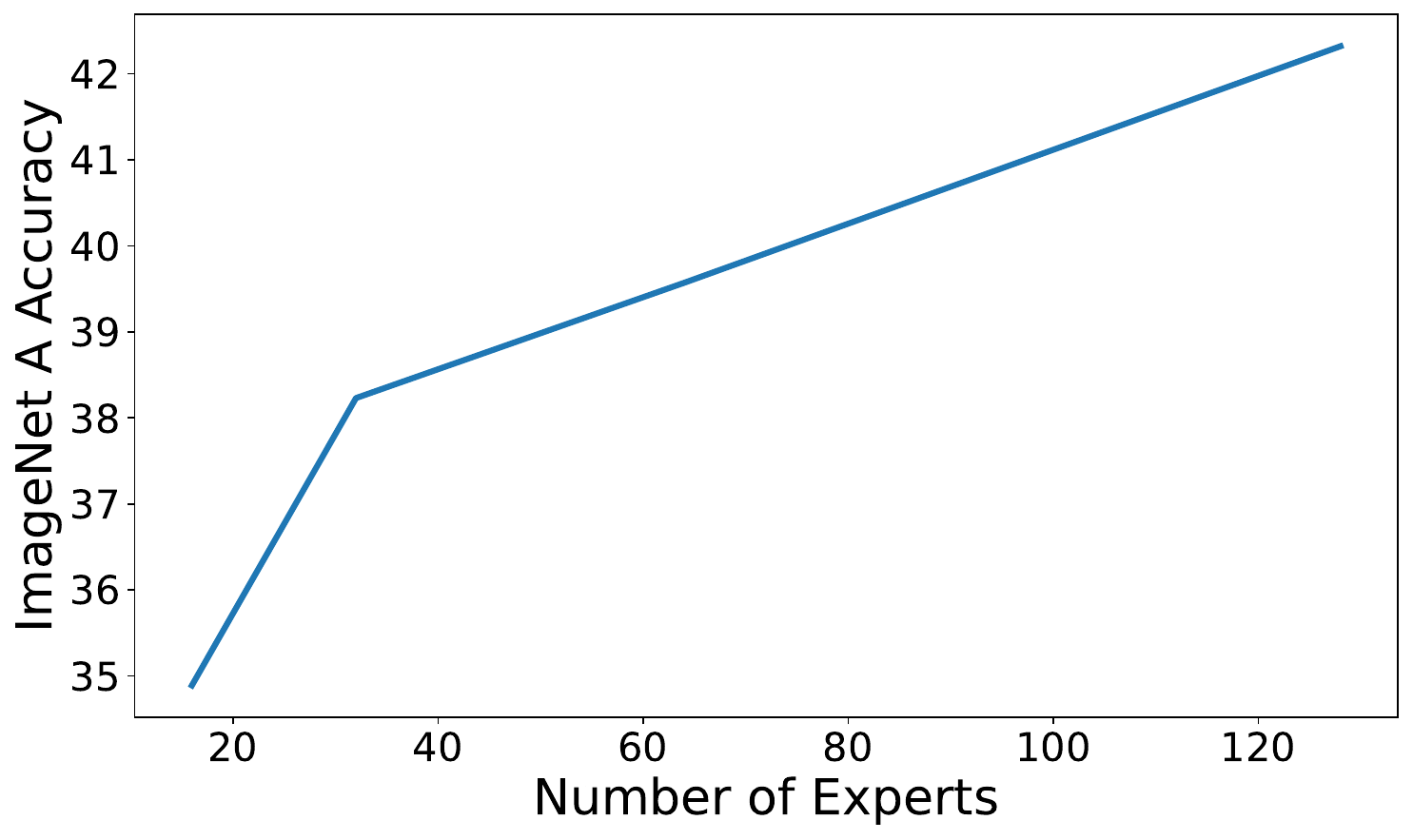}
    \caption{ImageNet A}
    \label{fig:sub2}
\end{subfigure}

\begin{subfigure}[b]{0.45\textwidth}
    \centering
    \includegraphics[width=\textwidth]{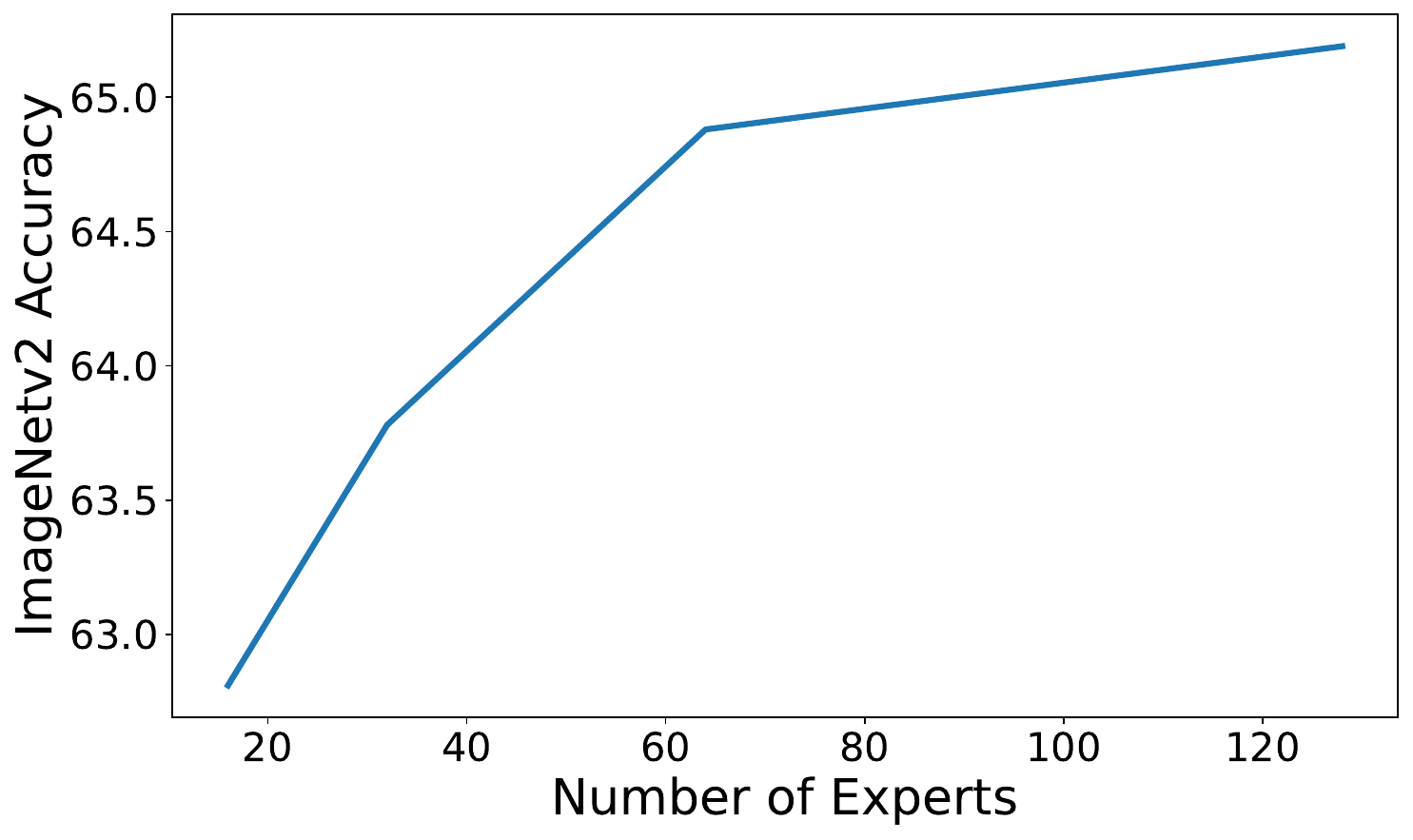}
    \caption{ImageNetv2}
    \label{fig:sub3}
\end{subfigure}
\hfill
\begin{subfigure}[b]{0.45\textwidth}
    \centering
    \includegraphics[width=\textwidth]{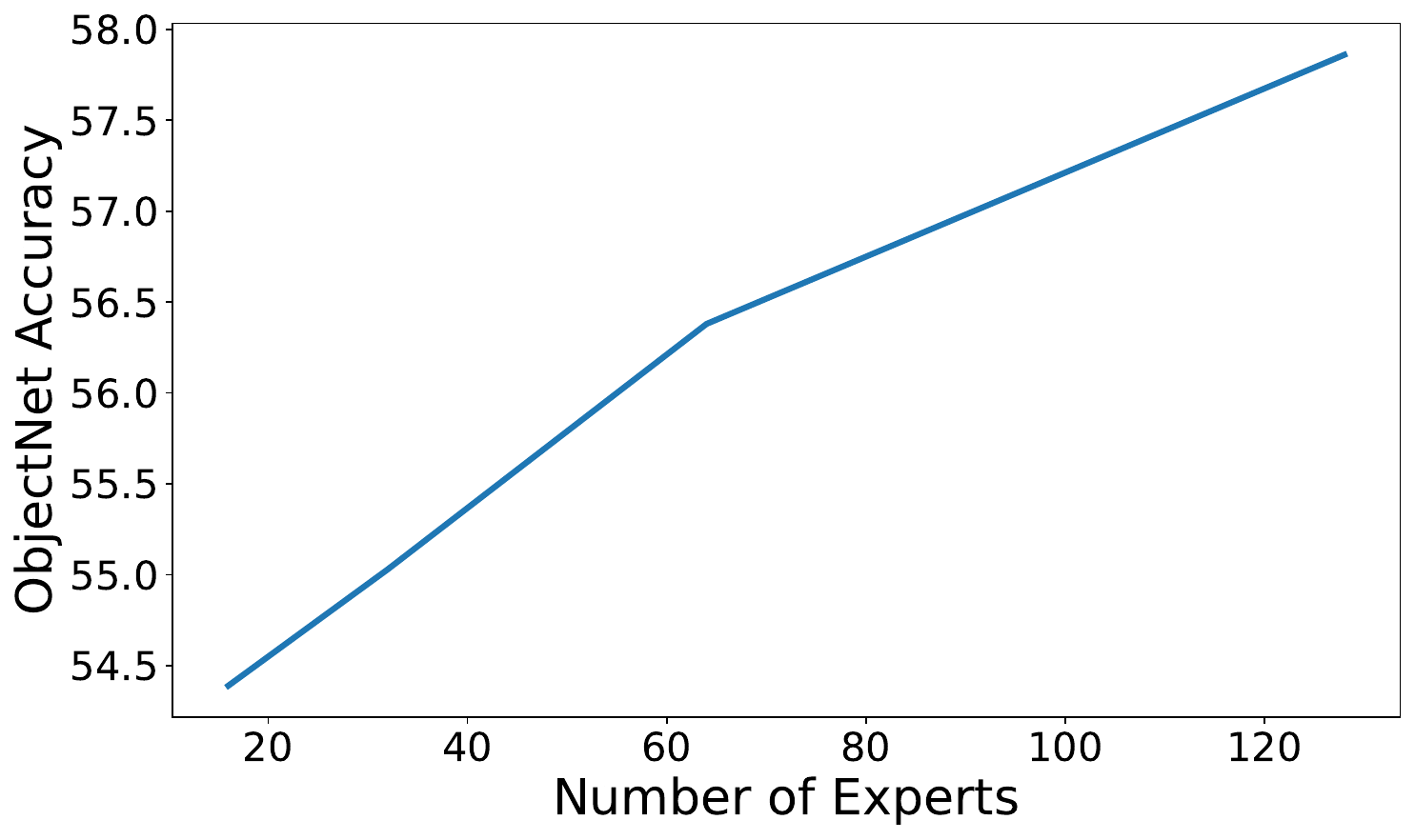}
    \caption{ObjectNet}
    \label{fig:sub4}
\end{subfigure}
\caption{Image classification accuracy by scaling the number of experts from 16 to 128, while keeping 8 activated experts.}
\label{scale_experts}
\end{figure}

Fig.~\ref{scale_experts} shows zero-shot performance on four image classification benchmarks: ImageNet-1K~\cite{imagenet1k}, ImageNetV2~\cite{imagenetv2}, ObjectNet~\cite{objectnet}, and ImageNet-A~\cite{imageneta}. As seen, increasing the total number of experts consistently improves accuracy on all benchmarks. Given our resource constraints, we adopt 32 total experts when scaling to larger vision encoders, which provides a favorable balance between performance and training cost.

\section{Implementation Details}

Table~\ref{arch} summarizes the architectures of our MoE-ViE models. Our implementation is based on OpenCLIP \cite{openclip}. Particularly, we remove the class token, as we did not observe any benefits out of it, and instead use the outputs from attention pooling. We adopt 2D RoPE \cite{rope} in the vision tower and absolute positional embedding in the text tower. We further replace standard FFN layers with SwiGLU \cite{glu} style, bringing the vision encoder design closer to that commonly used in LLM decoders. Below we describe the training and evaluation setup in detail.

\begin{table}[ht]
\centering
\caption{Configurations of MoE-ViE models.}
\label{arch}
\fontsize{6.5}{8}\selectfont
\resizebox{\textwidth}{!}{
\begin{tabular}{l|cc|cc|cc}
\hline
Scale & \multicolumn{2}{c|}{\textbf{B}} & \multicolumn{2}{c|}{\textbf{L}} & \multicolumn{2}{c}{\textbf{H}} \\
Encoder & \text{vision tower} & \text{text tower} & \text{vision tower} & \text{text tower} & \text{vision tower} & \text{text tower} \\
\hline\hline
\multirow{2}{*}{Params} & 0.46B & \multirow{2}{*}{0.30B} & 1.67B & \multirow{2}{*}{0.30B} & 3.50B & \multirow{2}{*}{0.81B} \\
 & (0.10B activated) & & (0.31B activated) & & (1.06B activated) & \\
Width & 768 & 1024 & 1024 & 1024 & 1280 & 1408 \\
Depth & 12 & 24 & 24 & 24 & 32 & 34 \\
\multirow{2}{*}{MLP} & 512 $\times$ 32 & \multirow{2}{*}{2730} & 684 $\times$ 32 & \multirow{2}{*}{2736} & 854 $\times$ 32 & \multirow{2}{*}{3754} \\
 & (4 activated) & & (4 activated) & & (8 activated) & \\
Heads & 12 & 16 & 16 & 16 & 16 & 16 \\
CLIP Dim & 1024 & 1024 & 1024 & 1024 & 1024 & 1024 \\
Pooling & Attn Pool & EOS Token & Attn Pool & EOS Token & Attn Pool & EOS Token \\
Context length & -- & 144 & -- & 144 & -- & 144 \\
Patch size & 16 & -- & 16 & -- & 14 & -- \\
Pos. embedding & RoPE & Absolute & RoPE & Absolute & RoPE & Absolute \\
\hline
\end{tabular}
}
\end{table}

\subsection{Training}
\label{appx_train_spec}

We perform contrastive pretraining of MoE-ViE models with progressively increasing image resolutions. Their schedules are as follows:
\begin{itemize}
\item MoE-ViE-B training includes image resolutions $112\times112$, $160\times160$, $224\times224$, 48.1B samples seen in total. 
\item MoE-ViE-L training includes image resolutions $112\times112$, $160\times160$, $224\times224$, $336\times336$, $384\times384$, 48.2B samples seen in total.
\item MoE-ViE-H training includes image resolutions $98\times98$,  $154\times154$, $224\times224$, $336\times336$, $448\times448$, 60.9B samples seen in total.
\end{itemize}
Our training data contains 2B image-text pairs from MetaCLIP \cite{metaclip} and an additional 1.5B proprietary data. We set learning rate to $10^{-3}$ with cosine decay. We use LAMB \cite{lamb} optimizer with weight decay 0.05 to support the large batch size 262144 being used. $(\beta_1,\beta_2)$ is set to $(0.9, 0.995)$. We reduce the batch size by $2\times$ at the highest resolution to save the compute cost. We only use InfoNCE loss \cite{infonce} during pretraining. 

For video finetuning, we set the learning rate to $10^{-6}$, and train for 20.5M samples for all MoE-ViE sizes. The training data consists of data curated by \cite{pe,merlot}. We use the same optimizer as in pretraining, with batch size of 4096. We uniformly sample 8 frames for each video input. The weight $\beta$ for distillation loss is set to 0.5. For alignment, Table~\ref{tab:training_stages} describes our training configurations in detail.

\subsection{Evaluation}
\label{appx_eval_spec}

To evaluate zero-shot classification and retrieval capability, we leverage CLIP benchmark framework \cite{clip_bench}. We use the same prompt templates as provided in \cite{pe}. Following \cite{eva-clip-18b}, we report the maximum score obtained under two data preprocessing methods, \ie, with and without center crop. We also perform reweighting \cite{internvl} on the retrieval results. 

We use the popular lmms-eval framework \cite{lmms_eval} to evaluate our alignment results. Greedy decoding is used in all cases.

\section{Additional Results}
\label{appx_additional_results}

\subsection{Linear Probing}
Following recent works \cite{pe,dino,internvl}, we extend our classification evaluation to include linear probing on the ImageNet-1K \cite{imagenet1k}. We show the results in Table~\ref{linear_probing_imagenet1k}, alongside other vision encoders which we benchmark. We see MoE-ViE-H/14 outperforms all encoders tested, even those with significantly more active parameters.

\begin{table}[htbp]
\centering
\caption{Linear probing results on ImageNet-1K.}
\label{linear_probing_imagenet1k}
\fontsize{6.5}{8}\selectfont
\setlength{\tabcolsep}{3pt}
\newcommand{\gc}{\cellcolor{gray!10}}

\begin{tabular}{l c c}
\toprule
\textbf{Model} & \textbf{Active Params} & \textbf{ImageNet-1K} \\
\midrule
DINOv2-g \cite{dinov2}        & 1.1B & 86.39 \\
SigLIP2-g-opt/16 \cite{siglip2} & 1.1B & 88.32 \\
PE\textsubscript{core}G/14 \cite{pe}        & 1.9B & 88.80 \\
InternViT-6B \cite{internvl}      & 5.5B & 88.44 \\
MoE-ViE-H/14    & 1.1B   & \textbf{88.85} \\
\bottomrule
\end{tabular}
\end{table}

\subsection{Alignment}
\label{appx_alignment_results}

To further investigate the potential of MoE-ViE in the alignment setting, we expand the training pipeline presented in Section~\ref{subsec:align}. We use Llama 3.1 8B Instruct \cite{llama3} as the LLM decoder. As detailed in Table~\ref{tab:training_stages}, we extend both the pre-alignment and supervised fine-tuning stages, scaling the total training data to 42M samples. Table~\ref{tab:model_performance_scaled} reports the results. MoE-ViE-H exhibits strong data scaling behavior: with 42M samples, it surpasses PE\textsubscript{lang}, which is mid-trained on 70M images and uses $1.7\times$ more active parameters, across all three evaluation axes, achieving 81.3 average on image benchmarks, 63.1 on video, and 132.6 on captioning. We posit that, as observed in pretraining, data quality and scale are the decisive factors for absolute alignment performance, and that MoE-ViE's sparse architecture is particularly well-suited to capitalize on these gains.

\begin{table}[htbp]
  \centering
  \caption{Training configuration overview across stages. Stage 2a is our 1k sample VE freeze step described in Section~\ref{subsec:align}.}
  \label{tab:training_stages}
  \fontsize{6.5}{8}\selectfont
  \setlength{\tabcolsep}{8pt} 
  \begin{tabular}{@{}lcccc@{}}
    \toprule
    & \textbf{S1} & \textbf{S2a} & \textbf{S2b} & \textbf{S3} \\
    \midrule
    Trainable Params & Projector & LLM + Proj. & All & All \\
    \midrule
    Steps            & 8,000  & 1,000  & 34,000 & 15,000 \\
    Global Batch Size& 2,048  & 512    & 512    & 512    \\
    Total Samples    & 16.4M  & 0.5M   & 17.4M  & 7.7M  \\
    Learning Rate    & $10^{-4}$ & $4{\times}10^{-5}$ & $4{\times}10^{-5}$ & $10^{-5}$ \\
    \midrule
    Max Seq. Length  & 1,280  & 6,144  & 6,144  & 11,520 \\
    Max Video Frames & 8      & 16     & 16     & 32     \\
    Max Num. Tiles   & 1      & 4      & 4      & 4      \\
    \bottomrule
  \end{tabular}
\end{table}

\begin{table}[htbp]
\centering
\caption{Data scaling experiments with MoE-ViE. Columns are grouped by task type. We use a maximum of 4 tiles.}
\label{tab:model_performance_scaled}
\fontsize{6.5}{8}\selectfont
\setlength{\tabcolsep}{3pt}
\newcommand{\gc}{\cellcolor{gray!15}}
\resizebox{\textwidth}{!}{%
\begin{tabular}{l c c c c c c c c c c c c c c c}
\toprule
& & & \multicolumn{6}{c}{\textbf{Image}} & \multicolumn{4}{c}{\textbf{Video}} & \multicolumn{3}{c}{\textbf{Captioning}} \\
\cmidrule(lr){4-9} \cmidrule(lr){10-13} \cmidrule(lr){14-16}
\textbf{Model}
& \rotatebox{90}{\shortstack{Training\\scale}}
& \rotatebox{90}{\shortstack{Active\\params}}
& \rotatebox{90}{Avg.\ Images}
& \rotatebox{90}{AI2D\cite{ai2d}}
& \rotatebox{90}{TextVQA\cite{textvqa}}
& \rotatebox{90}{ChartQA\cite{chartqa}}
& \rotatebox{90}{DocVQA\cite{docvqa}}
& \rotatebox{90}{Info.\ QA\cite{infovqa}}
& \rotatebox{90}{Avg.\ Video}
& \rotatebox{90}{VideoMME\cite{videomme}}
& \rotatebox{90}{MVBench\cite{mvbench}}
& \rotatebox{90}{EgoSchema\cite{egoschema}}
& \rotatebox{90}{Avg.\ Cap.}
& \rotatebox{90}{COCO\cite{coco}}
& \rotatebox{90}{NoCaps\cite{nocaps}} \\
\midrule
SigLIP2-g-opt/16 \cite{siglip2}              & --     & 1.1B & \gc 59.0 & 72.4 & 70.3 & 63.1 & 55.3 & 34.0 & \gc 49.5 & 46.2 & 48.5 & 53.8 & \gc 129.1 & 137.8 & 120.3 \\
PE\textsubscript{core}G/14 \cite{pe}         & --     & 1.9B & \gc 69.2 & 69.7 & 74.3 & 73.4 & 81.2 & 47.6 & \gc 48.6 & 46.0 & 48.7 & 51.2 & \gc 123.7 & 134.5 & 112.9 \\
InternViT2.5/14 \cite{internvl2p5}           & --     & 5.5B & \gc 68.1 & 72.9 & 71.3 & 74.6 & 74.3 & 47.6 & \gc 48.7 & 46.0 & 49.6 & 50.6 & \gc 123.0 & 132.5 & 113.5 \\
AIMv2 3B/14 \cite{aimv2}                      & --     & 2.7B & \gc 69.8 & 72.2 & 79.2 & 73.0 & 78.2 & 46.5 & \gc 49.7 & 49.6 & 49.9 & 49.6 & \gc 130.6 & 139.7 & 121.5 \\
\midrule
MoE-ViE-H/14                                  & 14M    & 1.1B & \gc 75.2 & 84.9 & 75.7 & 74.2 & 86.1 & 55.0 & \gc 58.9 & 52.0 & 63.6 & 61.2 & \gc 127.5 & 135.4 & 119.6 \\
PE\textsubscript{lang}G/14 \cite{pe}              & {>}70M & 1.9B & \gc 79.3 & 75.0 & 82.3 & 81.8 & 89.8 & 67.8 & \gc 54.1 & 49.6 & 52.6 & 60.0 & \gc 131.9 & 140.3 & 123.4 \\
MoE-ViE-H/14                                  & 42M    & 1.1B & \gc \textbf{81.3} & 90.5 & 79.9 & 80.8 & 90.4 & 64.8 & \gc \textbf{63.1} & 53.0 & 71.5 & 64.8 & \gc \textbf{132.6} & 141.5 & 123.6 \\
\bottomrule
\end{tabular}
}
\end{table}

\subsection{End-to-End VLM Latency}
\label{appx_e2e_latency}

In addition to the latency for vision encoder itself reported in Table~\ref{tab:performance_results}, we measure end-to-end inference latency after aligning the vision encoder to an LLM decoder (Llama~3.1 8B Instruct \cite{llama3}). Table~\ref{end2end_latency} compares different vision encoders in the resulting VLM. All of our benchmarking results on latency are conducted on a single H100 GPU. The overall trend matches Table~\ref{tab:performance_results} that MoE-ViE delivers comparable system-level latency while achieving substantially higher accuracy than the dense model with comparable size. Moreover, MoE-ViE attains significantly lower latency than the PE \cite{pe}, with comparable accuracy achieved. These results indicate that the efficiency benefits of MoE-ViE extend beyond the vision encoder and translate directly into improved latency vs. accuracy trade-offs in downstream VLM applications.

\begin{table}[htbp]
\centering
\caption{End-to-end latency with different vision encoders.}
\label{end2end_latency}
\fontsize{6.5}{8}\selectfont
\setlength{\tabcolsep}{6pt} 
\begin{tabular}{lccccc}
\toprule
Model & Total params & Active params & Batch size & \# of tiles & Latency (ms) \\
\midrule
SigLIP2-g-opt/16 \cite{siglip2}  & 1.1B  & 1.1B & 2 & 4   & 432.8  \\
PE\textsubscript{core}G/14 \cite{pe} & 1.9B  & 1.9B & 2 & 4  & 493.2  \\
MoE-ViE-H/14   & 3.5B  & 1.1B & 2 & 4   & 433.3  \\
\midrule
SigLIP2-g-opt/16 \cite{siglip2} &1.1B   & 1.1B & 2 & 16  & 523.4  \\
PE\textsubscript{core}G/14 \cite{pe} & 1.9B & 1.9B & 2 & 16  & 587.4  \\
MoE-ViE-H/14  & 3.5B  & 1.1B & 2 & 16  & 520.7  \\
\midrule
SigLIP2-g-opt/16 \cite{siglip2}  &1.1B  & 1.1B & 2 & 36  & 654.7  \\
PE\textsubscript{core}G/14 \cite{pe} &1.9B  & 1.9B & 2 & 36  & 744.0  \\
MoE-ViE-H/14  &3.5B   & 1.1B & 2 & 36  & 660.9  \\
\midrule
SigLIP2-g-opt/16 \cite{siglip2} & 1.1B   & 1.1B & 4 & 36  & 1297.9 \\
PE\textsubscript{core}G/14 \cite{pe} & 1.9B  & 1.9B & 4 & 36  & 1416.8 \\
MoE-ViE-H/14 & 3.5B    & 1.1B & 4 & 36  & 1300.9 \\
\bottomrule
\end{tabular}
\end{table}

%
%
\bibliographystyle{splncs04}
\bibliography{main}

\begin{thebibliography}{10}
\providecommand{\url}[1]{\texttt{#1}}
\providecommand{\urlprefix}{URL }
\providecommand{\doi}[1]{https://doi.org/#1}

\bibitem{gptoss}
Agarwal, S., Ahmad, L., Ai, J., Altman, S., Applebaum, A., Arbus, E., Arora,
  R.K., Bai, Y., Baker, B., Bao, H., et~al.: gpt-oss-120b \& gpt-oss-20b model
  card. arXiv preprint arXiv:2508.10925  (2025)

\bibitem{nocaps}
Agrawal, H., Desai, K., Wang, Y., Chen, X., Jain, R., Johnson, M., Batra, D.,
  Parikh, D., Lee, S., Anderson, P.: nocaps: novel object captioning at scale.
  In: Proceedings of the IEEE/CVF International Conference on Computer Vision
  (ICCV) (October 2019)

\bibitem{qwen3vl}
Bai, S., Cai, Y., Chen, R., Chen, K., Chen, X., Cheng, Z., Deng, L., Ding, W.,
  Gao, C., Ge, C., et~al.: Qwen3-vl technical report. arXiv preprint
  arXiv:2511.21631  (2025)

\bibitem{objectnet}
Barbu, A., Mayo, D., Alverio, J., Luo, W., Wang, C., Gutfreund, D., Tenenbaum,
  J., Katz, B.: Objectnet: A large-scale bias-controlled dataset for pushing
  the limits of object recognition models. Advances in neural information
  processing systems  \textbf{32} (2019)

\bibitem{pe}
Bolya, D., Huang, P.Y., Sun, P., Cho, J.H., Madotto, A., Wei, C., Ma, T., Zhi,
  J., Rajasegaran, J., Rasheed, H.A., Wang, J., Monteiro, M., Xu, H., Dong, S.,
  Ravi, N., Li, S.W., Dollar, P., Feichtenhofer, C.: Perception encoder: The
  best visual embeddings are not at the output of the network. In: The
  Thirty-ninth Annual Conference on Neural Information Processing Systems
  (2025), \url{https://openreview.net/forum?id=INqBOmwIpG}

\bibitem{food}
Bossard, L., Guillaumin, M., Van~Gool, L.: Food-101--mining discriminative
  components with random forests. In: European conference on computer vision.
  pp. 446--461. Springer (2014)

\bibitem{dino}
Caron, M., Touvron, H., Misra, I., J{\'e}gou, H., Mairal, J., Bojanowski, P.,
  Joulin, A.: Emerging properties in self-supervised vision transformers. In:
  Proceedings of the IEEE/CVF international conference on computer vision. pp.
  9650--9660 (2021)

\bibitem{cc12m}
Changpinyo, S., Sharma, P., Ding, N., Soricut, R.: Conceptual 12m: Pushing
  web-scale image-text pre-training to recognize long-tail visual concepts. In:
  Proceedings of the IEEE/CVF conference on computer vision and pattern
  recognition. pp. 3558--3568 (2021)

\bibitem{early_moe_1}
Chen, K., Xu, L., Chi, H.: Improved learning algorithms for mixture of experts
  in multiclass classification. Neural networks  \textbf{12}(9),  1229--1252
  (1999)

\bibitem{adamv}
Chen, T., Chen, X., Du, X., Rashwan, A., Yang, F., Chen, H., Wang, Z., Li, Y.:
  Adamv-moe: Adaptive multi-task vision mixture-of-experts. In: proceedings of
  the IEEE/CVF international conference on computer vision. pp. 17346--17357
  (2023)

\bibitem{pali}
Chen, X., Wang, X., Changpinyo, S., Piergiovanni, A.J., Padlewski, P., Salz,
  D., Goodman, S., Grycner, A., Mustafa, B., Beyer, L., et~al.: Pali: A
  jointly-scaled multilingual language-image model. arXiv preprint
  arXiv:2209.06794  (2022)

\bibitem{internvl2p5}
Chen, Z., Wang, W., Cao, Y., Liu, Y., Gao, Z., Cui, E., Zhu, J., Ye, S., Tian,
  H., Liu, Z., et~al.: Expanding performance boundaries of open-source
  multimodal models with model, data, and test-time scaling. arXiv preprint
  arXiv:2412.05271  (2024)

\bibitem{internvl}
Chen, Z., Wu, J., Wang, W., Su, W., Chen, G., Xing, S., Zhong, M., Zhang, Q.,
  Zhu, X., Lu, L., et~al.: Internvl: Scaling up vision foundation models and
  aligning for generic visual-linguistic tasks. In: Proceedings of the IEEE/CVF
  conference on computer vision and pattern recognition. pp. 24185--24198
  (2024)

\bibitem{clip_bench}
Cherti, M., Beaumont, R.: Clip benchmark (Nov 2022).
  \doi{10.5281/zenodo.15403103}, \url{https://doi.org/10.5281/zenodo.15403103}

\bibitem{openclip}
Cherti, M., Beaumont, R., Wightman, R., Wortsman, M., Ilharco, G., Gordon, C.,
  Schuhmann, C., Schmidt, L., Jitsev, J.: Reproducible scaling laws for
  contrastive language-image learning. In: Proceedings of the IEEE/CVF
  conference on computer vision and pattern recognition. pp. 2818--2829 (2023)

\bibitem{plm}
Cho, J.H., Madotto, A., Mavroudi, E., Afouras, T., Nagarajan, T., Maaz, M.,
  Song, Y., Ma, T., Hu, S., Jain, S., et~al.: Perceptionlm: Open-access data
  and models for detailed visual understanding. arXiv preprint arXiv:2504.13180
   (2025)

\bibitem{molmo2}
Clark, C., Zhang, J., Ma, Z., Park, J.S., Tripathi, R., Lee, S., Salehi, M.,
  Ren, J., Kim, C.D., Yang, Y., et~al.: Molmo2: Open weights and data for
  vision-language models with video understanding and grounding. In:
  Proceedings of the IEEE/CVF Conference on Computer Vision and Pattern
  Recognition. pp. 28652--28668 (2026)

\bibitem{gemini2p5}
Comanici, G., Bieber, E., Schaekermann, M., Pasupat, I., Sachdeva, N., Dhillon,
  I., Blistein, M., Ram, O., Zhang, D., Rosen, E., et~al.: Gemini 2.5: Pushing
  the frontier with advanced reasoning, multimodality, long context, and next
  generation agentic capabilities. arXiv preprint arXiv:2507.06261  (2025)

\bibitem{sigmoid2}
Csord{\'a}s, R., Irie, K., Schmidhuber, J.: Approximating two-layer feedforward
  networks for efficient transformers. In: Findings of the Association for
  Computational Linguistics: EMNLP 2023. pp. 674--692 (2023)

\bibitem{deepseekmoe}
Dai, D., Deng, C., Zhao, C., Xu, R., Gao, H., Chen, D., Li, J., Zeng, W., Yu,
  X., Wu, Y., et~al.: Deepseekmoe: Towards ultimate expert specialization in
  mixture-of-experts language models. In: Proceedings of the 62nd Annual
  Meeting of the Association for Computational Linguistics (Volume 1: Long
  Papers). pp. 1280--1297 (2024)

\bibitem{22bscale}
Dehghani, M., Djolonga, J., Mustafa, B., Padlewski, P., Heek, J., Gilmer, J.,
  Steiner, A.P., Caron, M., Geirhos, R., Alabdulmohsin, I., et~al.: Scaling
  vision transformers to 22 billion parameters. In: International conference on
  machine learning. pp. 7480--7512. PMLR (2023)

\bibitem{imagenet1k}
Deng, J., Dong, W., Socher, R., Li, L.J., Li, K., Fei-Fei, L.: Imagenet: A
  large-scale hierarchical image database. In: 2009 IEEE conference on computer
  vision and pattern recognition. pp. 248--255. Ieee (2009)

\bibitem{maskclip}
Dong, X., Bao, J., Zheng, Y., Zhang, T., Chen, D., Yang, H., Zeng, M., Zhang,
  W., Yuan, L., Chen, D., et~al.: Maskclip: Masked self-distillation advances
  contrastive language-image pretraining. In: Proceedings of the IEEE/CVF
  conference on computer vision and pattern recognition. pp. 10995--11005
  (2023)

\bibitem{attn_insights}
Dong, Y., Noci, L., Khodak, M., Li, M.: Attention retrieves, mlp memorizes:
  Disentangling trainable components in the transformer. arXiv e-prints pp.
  arXiv--2506 (2025)

\bibitem{switch}
Fedus, W., Zoph, B., Shazeer, N.: Switch transformers: Scaling to trillion
  parameter models with simple and efficient sparsity. Journal of Machine
  Learning Research  \textbf{23}(120),  1--39 (2022)

\bibitem{aimv2}
Fini, E., Shukor, M., Li, X., Dufter, P., Klein, M., Haldimann, D., Aitharaju,
  S., da~Costa, V.G.T., Béthune, L., Gan, Z., Toshev, A.T., Eichner, M., Nabi,
  M., Yang, Y., Susskind, J.M., El-Nouby, A.: Multimodal autoregressive
  pre-training of large vision encoders (2024),
  \url{https://arxiv.org/abs/2411.14402}

\bibitem{videomme}
Fu, C., Dai, Y., Luo, Y., Li, L., Ren, S., Zhang, R., Wang, Z., Zhou, C., Shen,
  Y., Zhang, M., Chen, P., Li, Y., Lin, S., Zhao, S., Li, K., Xu, T., Zheng,
  X., Chen, E., Shan, C., He, R., Sun, X.: Video-mme: The first-ever
  comprehensive evaluation benchmark of multi-modal llms in video analysis
  (2025), \url{https://arxiv.org/abs/2405.21075}

\bibitem{imagebind}
Girdhar, R., El-Nouby, A., Liu, Z., Singh, M., Alwala, K.V., Joulin, A., Misra,
  I.: Imagebind: One embedding space to bind them all. In: Proceedings of the
  IEEE/CVF conference on computer vision and pattern recognition. pp.
  15180--15190 (2023)

\bibitem{llama3}
Grattafiori, A., Dubey, A., Jauhri, A., Pandey, A., Kadian, A., Al-Dahle, A.,
  Letman, A., Mathur, A., Schelten, A., Vaughan, A., Yang, A., Fan, A., et~al.:
  The llama 3 herd of models (2024), \url{https://arxiv.org/abs/2407.21783}

\bibitem{treeoflife}
Gu, J., Stevens, S., Campolongo, E.G., Thompson, M.J., Zhang, N., Wu, J.,
  Kopanev, A., Mai, Z., White, A.E., Balhoff, J., et~al.: Bioclip 2: Emergent
  properties from scaling hierarchical contrastive learning. arXiv preprint
  arXiv:2505.23883  (2025)

\bibitem{seed1p5vl}
Guo, D., Wu, F., Zhu, F., Leng, F., Shi, G., Chen, H., Fan, H., Wang, J.,
  Jiang, J., Wang, J., et~al.: Seed1. 5-vl technical report. arXiv preprint
  arXiv:2505.07062  (2025)

\bibitem{imagenetr}
Hendrycks, D., Basart, S., Mu, N., Kadavath, S., Wang, F., Dorundo, E., Desai,
  R., Zhu, T., Parajuli, S., Guo, M., et~al.: The many faces of robustness: A
  critical analysis of out-of-distribution generalization. In: Proceedings of
  the IEEE/CVF international conference on computer vision. pp. 8340--8349
  (2021)

\bibitem{imageneta}
Hendrycks, D., Zhao, K., Basart, S., Steinhardt, J., Song, D.: Natural
  adversarial examples. In: Proceedings of the IEEE/CVF conference on computer
  vision and pattern recognition. pp. 15262--15271 (2021)

\bibitem{early_moe_2}
Jacobs, R.A., Jordan, M.I., Nowlan, S.J., Hinton, G.E.: Adaptive mixtures of
  local experts. Neural computation  \textbf{3}(1),  79--87 (1991)

\bibitem{align}
Jia, C., Yang, Y., Xia, Y., Chen, Y.T., Parekh, Z., Pham, H., Le, Q., Sung,
  Y.H., Li, Z., Duerig, T.: Scaling up visual and vision-language
  representation learning with noisy text supervision. In: International
  conference on machine learning. pp. 4904--4916. PMLR (2021)

\bibitem{mixtral}
Jiang, A.Q., Sablayrolles, A., Roux, A., Mensch, A., Savary, B., Bamford, C.,
  Chaplot, D.S., Casas, D.d.l., Hanna, E.B., Bressand, F., et~al.: Mixtral of
  experts. arXiv preprint arXiv:2401.04088  (2024)

\bibitem{early_moe_4}
Jordan, M.I., Jacobs, R.A.: Hierarchical mixtures of experts and the em
  algorithm. Neural computation  \textbf{6}(2),  181--214 (1994)

\bibitem{kinetics}
Kay, W., Carreira, J., Simonyan, K., Zhang, B., Hillier, C., Vijayanarasimhan,
  S., Viola, F., Green, T., Back, T., Natsev, P., et~al.: The kinetics human
  action video dataset. arXiv preprint arXiv:1705.06950  (2017)

\bibitem{ai2d}
Kembhavi, A., Salvato, M., Kolve, E., Seo, M., Hajishirzi, H., Farhadi, A.: A
  diagram is worth a dozen images. In: European conference on computer vision.
  pp. 235--251. Springer (2016)

\bibitem{cars}
Krause, J., Stark, M., Deng, J., Fei-Fei, L.: 3d object representations for
  fine-grained categorization. In: Proceedings of the IEEE international
  conference on computer vision workshops. pp. 554--561 (2013)

\bibitem{hmdb}
Kuehne, H., Jhuang, H., Garrote, E., Poggio, T., Serre, T.: Hmdb: a large video
  database for human motion recognition. In: 2011 International conference on
  computer vision. pp. 2556--2563. IEEE (2011)

\bibitem{gshard}
Lepikhin, D., Lee, H., Xu, Y., Chen, D., Firat, O., Huang, Y., Krikun, M.,
  Shazeer, N., Chen, Z.: Gshard: Scaling giant models with conditional
  computation and automatic sharding. arXiv preprint arXiv:2006.16668  (2020)

\bibitem{moe_cumo}
Li, J., Wang, X., Zhu, S., Kuo, C.W., Xu, L., Chen, F., Jain, J., Shi, H., Wen,
  L.: Cumo: Scaling multimodal llm with co-upcycled mixture-of-experts.
  Advances in Neural Information Processing Systems  \textbf{37},
  131224--131246 (2024)

\bibitem{mvbench}
Li, K., Wang, Y., He, Y., Li, Y., Wang, Y., Liu, Y., Wang, Z., Xu, J., Chen,
  G., Lou, P., Wang, L., Qiao, Y.: Mvbench: A comprehensive multi-modal video
  understanding benchmark. In: 2024 IEEE/CVF Conference on Computer Vision and
  Pattern Recognition (CVPR). pp. 22195--22206 (2024)

\bibitem{moe_llava}
Lin, B., Tang, Z., Ye, Y., Huang, J., Zhang, J., Pang, Y., Jin, P., Ning, M.,
  Luo, J., Yuan, L.: Moe-llava: Mixture of experts for large vision-language
  models. IEEE Transactions on Multimedia  (2026)

\bibitem{coco}
Lin, T.Y., Maire, M., Belongie, S., Hays, J., Perona, P., Ramanan, D.,
  Doll{\'a}r, P., Zitnick, C.L.: Microsoft coco: Common objects in context. In:
  European conference on computer vision. pp. 740--755. Springer (2014)

\bibitem{deepseekv3}
Liu, A., Feng, B., Xue, B., Wang, B., Wu, B., Lu, C., Zhao, C., Deng, C.,
  Zhang, C., Ruan, C., et~al.: Deepseek-v3 technical report. arXiv preprint
  arXiv:2412.19437  (2024)

\bibitem{llava_next}
Liu, H., Li, C., Li, Y., Lee, Y.J.: Improved baselines with visual instruction
  tuning. In: Proceedings of the IEEE/CVF conference on computer vision and
  pattern recognition. pp. 26296--26306 (2024)

\bibitem{aircraft}
Maji, S., Rahtu, E., Kannala, J., Blaschko, M., Vedaldi, A.: Fine-grained
  visual classification of aircraft. arXiv preprint arXiv:1306.5151  (2013)

\bibitem{egoschema}
Mangalam, K., Akshulakov, R., Malik, J.: Egoschema: A diagnostic benchmark for
  very long-form video language understanding. In: Advances in Neural
  Information Processing Systems (2023)

\bibitem{chartqa}
Masry, A., Long, D.X., Tan, J.Q., Joty, S., Hoque, E.: {C}hart{QA}: A benchmark
  for question answering about charts with visual and logical reasoning. In:
  Findings of the Association for Computational Linguistics: ACL 2022. pp.
  2263--2279. Association for Computational Linguistics (2022)

\bibitem{infovqa}
Mathew, M., Bagal, V., Tito, R.P., Karatzas, D., Valveny, E., Jawahar, C.V.:
  Infographicvqa (2021), \url{https://arxiv.org/abs/2104.12756}

\bibitem{docvqa}
Mathew, M., Karatzas, D., Jawahar, C.V.: Docvqa: A dataset for vqa on document
  images. In: 2021 IEEE Winter Conference on Applications of Computer Vision
  (WACV). pp. 2199--2208 (2021)

\bibitem{liemoe}
Mustafa, B., Riquelme, C., Puigcerver, J., Jenatton, R., Houlsby, N.:
  Multimodal contrastive learning with limoe: the language-image mixture of
  experts. Advances in Neural Information Processing Systems  \textbf{35},
  9564--9576 (2022)

\bibitem{sigmoid3}
Nguyen, H., Ho, N., Rinaldo, A.: Sigmoid gating is more sample efficient than
  softmax gating in mixture of experts. Advances in Neural Information
  Processing Systems  \textbf{37},  118357--118388 (2024)

\bibitem{flowers}
Nilsback, M.E., Zisserman, A.: Automated flower classification over a large
  number of classes. In: 2008 Sixth Indian conference on computer vision,
  graphics \& image processing. pp. 722--729. IEEE (2008)

\bibitem{infonce}
Oord, A.v.d., Li, Y., Vinyals, O.: Representation learning with contrastive
  predictive coding. arXiv preprint arXiv:1807.03748  (2018)

\bibitem{dinov2}
Oquab, M., Darcet, T., Moutakanni, T., Vo, H., Szafraniec, M., Khalidov, V.,
  Fernandez, P., Haziza, D., Massa, F., El-Nouby, A., Assran, M., Ballas, N.,
  Galuba, W., Howes, R.: Dinov2: Learning robust visual features without
  supervision. CoRR  \textbf{abs/2304.07193} (2023),
  \url{https://doi.org/10.48550/arXiv.2304.07193}

\bibitem{clip}
Radford, A., Kim, J.W., Hallacy, C., Ramesh, A., Goh, G., Agarwal, S., Sastry,
  G., Askell, A., Mishkin, P., Clark, J., et~al.: Learning transferable visual
  models from natural language supervision. In: International conference on
  machine learning. pp. 8748--8763. PmLR (2021)

\bibitem{imagenetv2}
Recht, B., Roelofs, R., Schmidt, L., Shankar, V.: Do imagenet classifiers
  generalize to imagenet? In: International conference on machine learning. pp.
  5389--5400. PMLR (2019)

\bibitem{vmoe}
Riquelme, C., Puigcerver, J., Mustafa, B., Neumann, M., Jenatton, R.,
  Susano~Pinto, A., Keysers, D., Houlsby, N.: Scaling vision with sparse
  mixture of experts. Advances in Neural Information Processing Systems
  \textbf{34},  8583--8595 (2021)

\bibitem{glu}
Shazeer, N.: Glu variants improve transformer. arXiv preprint arXiv:2002.05202
  (2020)

\bibitem{original_moe}
Shazeer, N., Mirhoseini, A., Maziarz, K., Davis, A., Le, Q., Hinton, G., Dean,
  J.: Outrageously large neural networks: The sparsely-gated mixture-of-experts
  layer. arXiv preprint arXiv:1701.06538  (2017)

\bibitem{mome}
Shen, L., Chen, G., Shao, R., Guan, W., Nie, L.: Mo{ME}: Mixture of multimodal
  experts for generalist multimodal large language models. In: The
  Thirty-eighth Annual Conference on Neural Information Processing Systems
  (2024), \url{https://openreview.net/forum?id=Xskl7Da34U}

\bibitem{textcap}
Sidorov, O., Hu, R., Rohrbach, M., Singh, A.: Textcaps: a dataset for image
  captioning with reading comprehension. In: European conference on computer
  vision. pp. 742--758. Springer (2020)

\bibitem{dinov3}
Siméoni, O., Vo, H.V., Seitzer, M., Baldassarre, F., Oquab, M., Jose, C.,
  Khalidov, V., Szafraniec, M., Yi, S., et~al.: Dinov3 (2025),
  \url{https://arxiv.org/abs/2508.10104}

\bibitem{textvqa}
Singh, A., Pang, G., et~al.: Textvqa: Towards reading and reasoning in visual
  question answering. In: Proceedings of the IEEE/CVF Conference on Computer
  Vision and Pattern Recognition (2019)

\bibitem{ucf}
Soomro, K., Zamir, A.R., Shah, M.: Ucf101: A dataset of 101 human actions
  classes from videos in the wild. arXiv preprint arXiv:1212.0402  (2012)

\bibitem{rope}
Su, J., Ahmed, M., Lu, Y., Pan, S., Bo, W., Liu, Y.: Roformer: Enhanced
  transformer with rotary position embedding. Neurocomputing  \textbf{568},
  127063 (2024)

\bibitem{eva-clip-18b}
Sun, Q., Wang, J., Yu, Q., Cui, Y., Zhang, F., Zhang, X., Wang, X.:
  Eva-clip-18b: Scaling clip to 18 billion parameters. arXiv preprint
  arXiv:2402.04252  (2024)

\bibitem{kimi2p5}
Team, K., Bai, T., Bai, Y., Bao, Y., Cai, S., Cao, Y., Charles, Y., Che, H.,
  Chen, C., Chen, G., et~al.: Kimi k2. 5: Visual agentic intelligence. arXiv
  preprint arXiv:2602.02276  (2026)

\bibitem{kimik2}
Team, K., Bai, Y., Bao, Y., Chen, G., Chen, J., Chen, N., Chen, R., Chen, Y.,
  Chen, Y., Chen, Y., et~al.: Kimi k2: Open agentic intelligence. arXiv
  preprint arXiv:2507.20534  (2025)

\bibitem{country}
Thomee, B., Shamma, D.A., Friedland, G., Elizalde, B., Ni, K., Poland, D.,
  Borth, D., Li, L.J.: Yfcc100m: The new data in multimedia research.
  Communications of the ACM  \textbf{59}(2),  64--73 (2016)

\bibitem{triton}
Tillet, P., Kung, H.T., Cox, D.: Triton: an intermediate language and compiler
  for tiled neural network computations. In: Proceedings of the 3rd ACM SIGPLAN
  International Workshop on Machine Learning and Programming Languages. pp.
  10--19 (2019)

\bibitem{siglip2}
Tschannen, M., Gritsenko, A., Wang, X., Naeem, M.F., Alabdulmohsin, I.,
  Parthasarathy, N., Evans, T., Beyer, L., Xia, Y., Mustafa, B., et~al.: Siglip
  2: Multilingual vision-language encoders with improved semantic
  understanding, localization, and dense features. arXiv preprint
  arXiv:2502.14786  (2025)

\bibitem{locca}
Wan, B., Tschannen, M., Xian, Y., Pavetic, F., Alabdulmohsin, I.M., Wang, X.,
  Susano~Pinto, A., Steiner, A., Beyer, L., Zhai, X.: Locca: Visual pretraining
  with location-aware captioners. Advances in Neural Information Processing
  Systems  \textbf{37},  116355--116387 (2024)

\bibitem{aux_free}
Wang, L., Gao, H., Zhao, C., Sun, X., Dai, D.: Auxiliary-loss-free load
  balancing strategy for mixture-of-experts. arXiv preprint arXiv:2408.15664
  (2024)

\bibitem{internvl3}
Wang, W., Gao, Z., Gu, L., Pu, H., Cui, L., Wei, X., Liu, Z., Jing, L., Ye, S.,
  Shao, J., et~al.: Internvl3. 5: Advancing open-source multimodal models in
  versatility, reasoning, and efficiency. arXiv preprint arXiv:2508.18265
  (2025)

\bibitem{moe_clipup}
Wang, X., Chen, C., Yang, Y., Chen, H.Y., Zhang, B., Pal, A., Zhu, X., Du, X.:
  Clip-up: A simple and efficient mixture-of-experts clip training recipe with
  sparse upcycling. arXiv preprint arXiv:2502.00965  (2025)

\bibitem{omni_smola}
Wu, J., Hu, X., Wang, Y., Pang, B., Soricut, R.: Omni-smola: Boosting
  generalist multimodal models with soft mixture of low-rank experts. In:
  Proceedings of the IEEE/CVF Conference on Computer Vision and Pattern
  Recognition. pp. 14205--14215 (2024)

\bibitem{rmoe}
Wu, L., Liu, M., Chen, Y., Chen, D., Dai, X., Yuan, L.: Residual mixture of
  experts. arXiv preprint arXiv:2204.09636  (2022)

\bibitem{tove}
Wu, Y., Du, J., Yan, K., Ding, S., Li, X.: To{VE}: Efficient vision-language
  learning via knowledge transfer from vision experts. In: The Thirteenth
  International Conference on Learning Representations (2025),
  \url{https://openreview.net/forum?id=EMMnAd3apQ}

\bibitem{metaclip}
Xu, H., Xie, S., Tan, X.E., Huang, P.Y., Howes, R., Sharma, V., Li, S.W.,
  Ghosh, G., Zettlemoyer, L., Feichtenhofer, C.: Demystifying clip data. arXiv
  preprint arXiv:2309.16671  (2023)

\bibitem{msr_vtt}
Xu, J., Mei, T., Yao, T., Rui, Y.: Msr-vtt: A large video description dataset
  for bridging video and language. In: Proceedings of the IEEE conference on
  computer vision and pattern recognition. pp. 5288--5296 (2016)

\bibitem{qwen3}
Yang, A., Li, A., Yang, B., Zhang, B., Hui, B., Zheng, B., Yu, B., Gao, C.,
  Huang, C., Lv, C., et~al.: Qwen3 technical report. arXiv preprint
  arXiv:2505.09388  (2025)

\bibitem{omnivinci}
Ye, H., Yang, C.H.H., Goel, A., Huang, W., Zhu, L., Su, Y., Lin, S., Cheng,
  A.C., Wan, Z., Tian, J., et~al.: Omnivinci: Enhancing architecture and data
  for omni-modal understanding llm. arXiv preprint arXiv:2510.15870  (2025)

\bibitem{lamb}
You, Y., Li, J., Reddi, S., Hseu, J., Kumar, S., Bhojanapalli, S., Song, X.,
  Demmel, J., Keutzer, K., Hsieh, C.J.: Large batch optimization for deep
  learning: Training bert in 76 minutes. In: International Conference on
  Learning Representations (2020),
  \url{https://openreview.net/forum?id=Syx4wnEtvH}

\bibitem{flickr}
Young, P., Lai, A., Hodosh, M., Hockenmaier, J.: From image descriptions to
  visual denotations: New similarity metrics for semantic inference over event
  descriptions. Transactions of the association for computational linguistics
  \textbf{2},  67--78 (2014)

\bibitem{cocca}
Yu, J., Wang, Z., Vasudevan, V., Yeung, L., Seyedhosseini, M., Wu, Y.: Coca:
  Contrastive captioners are image-text foundation models. arXiv preprint
  arXiv:2205.01917  (2022)

\bibitem{merlot}
Zellers, R., Lu, X., Hessel, J., Yu, Y., Park, J.S., Cao, J., Farhadi, A.,
  Choi, Y.: Merlot: Multimodal neural script knowledge models. Advances in
  neural information processing systems  \textbf{34},  23634--23651 (2021)

\bibitem{siglip}
Zhai, X., Mustafa, B., Kolesnikov, A., Beyer, L.: Sigmoid loss for language
  image pre-training. In: Proceedings of the IEEE/CVF international conference
  on computer vision. pp. 11975--11986 (2023)

\bibitem{clip_moe}
Zhang, J., Qu, X., Zhu, T., Cheng, Y.: Clip-moe: Towards building mixture of
  experts for clip with diversified multiplet upcycling. In: Proceedings of the
  2025 Conference on Empirical Methods in Natural Language Processing. pp.
  5406--5419 (2025)

\bibitem{lmms_eval}
Zhang, K., Li, B., Zhang, P., Pu, F., Cahyono, J.A., Hu, K., Liu, S., Zhang,
  Y., Yang, J., Li, C., Liu, Z.: Lmms-eval: Reality check on the evaluation of
  large multimodal models (2024), \url{https://arxiv.org/abs/2407.12772}

\end{thebibliography}
\end{document}